\pdfoutput=1
\documentclass[11pt]{article}

\usepackage[final]{acl}
\usepackage{subcaption}
\usepackage{times}
\usepackage{latexsym}
\usepackage{float}
\usepackage{xcolor}
\usepackage[inkscapelatex=false]{svg}
\usepackage{booktabs}
\usepackage{amsmath} 
\usepackage{tcolorbox}
\usepackage{CJKutf8}
\usepackage{comment}
\newcommand{\zh}[1]{\begin{CJK*}{UTF8}{gbsn}#1\end{CJK*}}
\newcommand{\ja}[1]{\begin{CJK*}{UTF8}{min}#1\end{CJK*}}
\usepackage[T1]{fontenc}
\usepackage[utf8]{inputenc}
\usepackage{array} % required for text wrapping in tables
\usepackage{microtype}
\usepackage{inconsolata}

\usepackage{graphicx}
\usepackage{adjustbox}

\title{Internal, Verbalised, and Perceived Confidence in
Multilingual LLMs}

 \author{Alessandra Terranova, Alexandra Birch, Björn Ross \\
         School of Informatics, University of Edinburgh \\
           \small{
    \textbf{Correspondence:} \href{mailto:email@domain}{a.terranova@ed.ac.uk}
  }}

\begin{document}
\maketitle

\begin{abstract}
As large language models (LLMs) are increasingly used by humans across multiple languages, there is a need for models to express their uncertainty in a way that is well-calibrated, honest, and aligned to how users interpret it.
   We construct the first multilingual and distributional dataset of LLM linguistic expressions of confidence, spanning seven languages. Each expression is labelled with a corresponding probability distribution and the belief intervals labelled by native-speaker annotators. We argue that human understanding of linguistic hedges and expressions of confidence varies cross-linguistically, and it is fuzzy and at times vague, which makes the use of point estimates for confidence insufficient. 
   We study expressions of confidence in LLMs both as linguistic and numerical verbalisations, and as internal token-probability and and consistency-based signals.
   We find that, in the multiple-choice-question setting, across elicitation methods, English is better calibrated than the other languages studied, and that models lose calibration when attempting to surface uncertainty, with logit extraction achieving the best calibration, followed by consistency-based methods and linguistic confidence, while verbalised numerical confidence exhibits the poorest calibration. Additionally, instruction tuning enhances discrimination but degrades calibration for internal methods, whereas it consistently improves calibration for verbalisation-based methods.
\end{abstract}

%The report should be no longer than 7000 words (excluding
%references), but the template and format is otherwise at the students
%discretion. We recommend the student selects a conference venue or
%journal relevant to the student’s field or the topical focus of their IEP
%project, and use whatever template they require. 

\section{Introduction}

Large language models (LLMs) are deployed globally, and increasingly used in high-stakes applications;
nonetheless, they are prone to hallucinations and factual errors, and often provide overconfident responses that may misplace user trust and lead to overreliance on confident but incorrect responses. Effective uncertainty communication is therefore essential for safe and reliable deployment.

A variety of methods have been proposed to extract uncertainty from large language models:
token probability-based \cite{duan2024shifting} and internal state-based methods \cite{kossen2024semantic, ji2025calibrating} typically require white-box access; black-box methods like consistency-based ones \cite{kuhn2023semantic, vazhentsev2023hybrid} require multiple generations, while verbalised numerical expressions of uncertainty  are efficient but not user-friendly \cite{tian2023just, xiong2023can, mielke2022reducing}.
Communicating uncertainty through natural language hedges \cite{yona2024largelanguagemodelsfaithfully, steyvers2025large, kim2024m} is a popular, human-centered approach, but current methods suffer from two critical limitations. First, existing research heavily focuses on English, largely ignoring how the usage and interpretation of epistemic markers differ across languages. 
Additionally, existing approaches that map hedging words to numerical scores (e.g., "probably" → 0.5, "likely" → 0.7) incorrectly treat hedges as precise point values rather than the fuzzy ranges humans actually perceive. Different speakers interpret the same hedge with considerable variance, and this variability may be further amplified across languages and cultures \cite{ fraser2010pragmatic, kranich2011hedge, uysal2013cross}.
We argue that encoding epistemic markers of uncertainty as probability distributions rather than fixed numbers not only better mimics human cognition but also enables systems to account for inter-speaker variability and language-specific interpretation patterns. 

The goal of this research is to \textit{create a multilingual framework that maps linguistic uncertainty to probability distributions, ensuring that AI systems use hedging expressions in a way that is calibrated to their internal confidence and better aligned with human interpretation.} In particular, we address the following research questions:

\begin{itemize}
    \item \textbf{RQ1.} Do the confidence ranges humans assign to hedging words match the internal confidence of LLMs and the confidence scores that LLMs verbalise?
    \item \textbf{RQ2.} How does linguistic uncertainty expression compare to verbalised numerical uncertainty, consistency, and token-probability based methods?
    \item \textbf{RQ3.} How does the uncertainty range of linguistic hedges vary across different languages, and do LLMs capture cross-linguistic nuances?
\end{itemize}

To explore these questions, we construct a multilingual benchmark of factual questions with answers containing hedging expressions that show varying degrees of certainty. Each hedging expression is mapped to a human-annotated probability distribution. To enable scalability, we also train a mapper from natural language expressions to probability distributions. Through this benchmark, we evaluate how multilingual LLMs express uncertainty, assess their calibration, and examine cross-linguistic variation.
Our main contributions are:

\begin{enumerate}

\item \textbf{A multilingual linguistic confidence benchmark (RQ3)} comprising human-annotated probability distributions for hedging expressions used by LLMs across seven typologically diverse languages (English, French, German, Chinese, Italian, Japanese, and Hindi).
\item \textbf{Comprehensive evaluation (RQ1, RQ2)} showing that models lose calibration when attempting to surface uncertainty: when answering multiple-choice questions, logit extraction achieves the best calibration, followed by multisampling and linguistic confidence methods. Verbalised numerical confidence (VNC) exhibits the poorest calibration. 
\item \textbf{Cross-linguistic analysis (RQ3)} showing a systematic performance gap between English and other languages across methods.
\item \textbf{Instruction-tuning effects (RQ2)}: For logits-based and multisampling methods, instruction tuning enhances discrimination and reduces error while degrading calibration. For verbalisation-based methods, instruction tuning consistently improves calibration.
\end{enumerate}

We release our code\footnote{GitHub: \href{https://github.com/terrales/multilingual_uncertainty}{multilingual\_uncertainty}} and dataset\footnote{HuggingFace: \href{https://huggingface.co/datasets/terrales02/multilingual_uncertainty}{multilingual\_uncertainty}}.
We hope this work establishes a foundation for linguistically grounded, calibrated uncertainty communication in multilingual AI systems, enabling safer deployment in high-stakes applications.

\section{Background}

\subsection{From the Linguistic Theory of Hedging to LLMs}
To investigate how language models express uncertainty, it is important to ground the analysis in the linguistic understanding of semantic fuzziness and hedging in pragmatics. George Lakoff's foundational work on hedges \cite{lakoff1973hedges} challenged the traditional view that sentences in natural language are either true or false. He argued that natural language concepts have vague boundaries, therefore category membership is a matter of degree instead of strictly binary, and one of the functions of hedges is to make things "fuzzier" or "less fuzzy". Building on this, Fraser notes \cite{fraser2010pragmatic} that hedging expressions are fundamental for pragmatic competence, as they allow speakers to indicate a modulated degree of commitment to the truth of a statement. %, to facilitate politeness, face-saving, and to mitigate face-threats in social interaction.

Terms like `I think' or `probably' are highly context-dependent; their usage by LLMs serves as a tool for epistemic sincerity, but models' verbalisations are frequently miscalibrated and do not reliably mirror the model's internal confidence \cite{yona2024largelanguagemodelsfaithfully}. This lack of pragmatic competence can lead to inaccurate information being communicated fluently and persuasively, and increase the risk of human overreliance \cite{steyvers2025large}. \citet{ulmer2026anthropomimetic} argue for \textit{anthropomimetic} uncertainty, showing that LLMs struggle to capture the pragmatic nuance of epistemic markers \cite{belem2024perceptions}. While epistemic markers surface frequently in generation, they fail to reliably track underlying model certainty \cite{liu-etal-2025-revisiting} , and additionally they can be misaligned to human expectations \cite{tang2026evaluation}.
Recent work by \citet{yona2026hallucinations} argues that user trust can be built on imperfect knowledge, as long as uncertainty is communicated honestly. %, and this awareness of uncertainty becomes even more important for enabling effective tool use in agentic systems.

\subsection{LLM Calibration}

Existing methods for confidence estimation and uncertainty communication face different barriers: approaches based on white-box token probabilities \cite{duan2024shifting, kuhn2023semantic} are  simple and efficient, but these methods require access to model logits, often unavailable for frontier commercial APIs. Recent work has explored auxiliary calibrators, training or using auxilary models to evaluate black-box model outputs without full internal logit access \cite{ulmer2024calibrating, li2024graph, xia2025influences}.
A second line of research estimates confidence through multiple generations \cite{vazhentsev2023hybrid, kuhn2023semantic}, but multisampling quickly becomes computationally expensive. A third approach directly elicits numerical probabilities by prompting the model to output a confidence score alongside its answer \cite{xiong2023can, lin2022teaching, mielke2022reducing}, but such explicit numerical outputs rarely align with natural human communication patterns.  
The post-training and alignment process itself is an additional barrier to to calibration: while pre-trained base models are typically well-calibrated, instruction-tuning can degrade this property, as reward models tend to favour high confidence \cite{tian2023just}. 

Using linguistic expressions of uncertainty through natural language hedges has emerged as a light-weight, human-centered method \cite{yona2024largelanguagemodelsfaithfully, steyvers2025large, kim2024m, mielke2022reducing}, but might still lead users to overestimate model confidence \cite{steyvers2025large, sun2025explainingsourcesuncertaintyautomated}.
\citet{tao2025revisiting} report that linguistic verbalised expressions of confidence can often outperform the numerical verbalisation methods, and -- in some cases -- token-probability proxies for black-box, single-pass settings.
We argue that the uncertainty of linguistic expressions should be modelled as a distribution, rather than as a single numerical confidence score. This has only been researched for few English hedges in restricted domains \cite{wang2025calibratingexpressionscertainty, illinoisPerceptionProbability}, but it may provide a more faithful and generalizable representation \cite{tao2025largelanguagemodelsexpress}.

\subsection{Cross-linguistic Variation}

The vast majority of current research on confidence and expressions of uncertainty in LLMs only targets English, but the usage and interpretation of linguistic hedges varies widely cross-linguistically and across cultures. 
Linguistic typology confirms language-specific norms with respects to the frequency and strength of hedges, and to the mechanisms through which they are communicated: Japanese employs weakeners to avoid over-assertiveness, often through pragmatic sentence-final particles \cite{itani1995semantics}; speakers of German, Mandarin, and Hindi tend to use more strengtheners \cite{kranich2011hedge, uysal2013cross}. English and French, instead, tend to use moderate expressions of confidence through modal verbs and adverbs \cite{celle2004french, dieuleveut2025interpreting}. In Italian the strength and prevalence of hedges seems to be highly context dependent \cite{donadio2022hedges, dheskali2017hedging}.

Limited attention has been devoted to multilingual expressions of confidence in LLMs.
\citet{krause2023confidently} show that GPT-3.5 is overconfident across languages, and \citet{huang2026investigating} look at the shortcomings of supervised fine-tuning for multilingual calibration. \citet{xue2025mlingconf} look at confidence in multilingual LLMs, but do not investigate linguistic expressions.
\citet{rathi2025humansoverrelyoverconfidentlanguage} argue that reliance on AI also varies significantly by culture. %A single, English-centric mapping of uncertainty fails to account for these different baselines, risking safety in global deployment.

\section{Building a Confidence Dataset}
To study user understanding of LLM uncertainty expressions, we first collect LLM-generated hedges across seven different languages; next, we obtain human annotations of confidence for a sample of the expressions through Prolific; finally, we use the intervals provided by annotators to fit confidence distributions for each expression. %The process is presented in Figure \ref{fig:dataset}, and 
Dataset construction is detailed in Appendix  \ref{sec:annotation}.

%\begin{figure}
%    \centering
%    \includegraphics[width=0.5\linewidth]{todo}
%    \caption{{\color{red}Overview of the dataset creation pipeline. The process involves generating candidate expressions using LLMs, validating them, collecting human interval annotations, filtering for quality, and fitting Beta distributions to represent linguistic confidence.}}
%    \label{fig:dataset}
%\end{figure}

\subsection{Dataset Selection}
We restrict our benchmark selection to short factual question-answering data. For English \citet{tao2025largelanguagemodelsexpress} annotated a subset of SimpleQA \cite{wei2024measuring} using point estimates. \citet{rathi2025humansoverrelyoverconfidentlanguage} instead explored LLM overconfidence in five languages through a machine-translated intersection of MMLU \cite{hendrycks2020measuring} and JMMLU, but did not curate a dataset. \citet{wang2025calibratingexpressionscertainty} and \citet{illinoisPerceptionProbability} have previously treated English hedges as probability distributions, but only in restricted domains and for a fixed set of 15-20 of expressions.

We use a subset of the culturally agnostic questions in GlobalMMLU Lite \cite{singh2025global} in English, French, German, Japanese, Mandarin, Hindi, and Italian. This data provides factual multiple-choice questions (MCQ), manually annotated and professionally translated, and labelled as not requiring cultural, regional, or dialect knowledge to be answered. 

\subsection{Data Augmentation}
We sample the same 100 questions for each of the seven languages selected and, following the prompt by  \citet{tao2025largelanguagemodelsexpress}, we use GPT-5 mini to expand each question with 50 different wordings of the correct answer, each containing a different expression of confidence for a lowest, low, moderate, and high confidence level
We deliberately use a commercial LLM to generate expressions of confidence because we aim to capture the linguistic artifacts that currently drive users' perception of AI uncertainty when interacting with AI "in the wild". 

\subsection{Validation}
We validate the augmented data through string comparison by ensuring that each of the sampled answers uses a different wording. Additionally, we manually check the quality of 100 samples in English and Italian. For other languages, we back-translate the answers into English as a first manual validation step. 
To aid validation, we use an LLM-as-a-judge (GPT-4-mini, prompt in Appendix \ref{sec:annotation}). Additionally, we compare the English expressions to those in Hyland's hedge terms (\citeyear{Hyland1998HedgingIS}) and find that 67.5\% contain at least one term, mostly everyday modal and evidential words.
We sample 80 question-answer pairs per language (20 randomly sampled questions × 4 balanced confidence-varying answers) for annotation, totalling 560 expressions across seven languages. 

\subsection{Data Annotation}
%The novelty of this framework lies in the transition from point estimates to distributional annotation. 
We fit distributions to capture the inherent fuzziness and asymmetry of linguistic uncertainty expressions \cite{ lakoff1973hedges, Zadeh1972-ZADAFS}. Beta distributions are appropriate because of their bounded [0, 1] domain and flexible shape parameters, nonetheless they assume that human nuance can be condensed in two parameters.

%Given the infeasibility of collecting sufficient point estimates to reliably fit distributions, we task annotators with providing belief intervals representing their perceived confidence range for each expression.
We recruit 10 crowdworkers per language on Prolific to annotate lower and upper confidence bounds for how likely the AI model appears to consider the given answer correct. Annotators additionally rate fluency and readability on 5-point scales as a further quality check, and, for languages other than English, complete a validated AI attitudes survey \cite{Schepman06012026}. 

To ensure data quality, we exclude unreliable annotators who fail our attention checks by assigning minimum confidence below 40\% to high-certainty expressions or maximum confidence above 60\% to low-certainty expressions. After this process, we retain 10/10 annotations in German, English, French, and Italian, 9/10 in Japanese, 8/10 in Chinese, and 7/10 in Hindi. The annotator intervals retained still contain real disagreement and disjoint belief intervals (Error analysis in Appendix \ref{sec:data_val}). We preserve inter-annotator variance when present, as this reflects the genuine ambiguity of hedging expressions. After filtering, each data point retains 7-10 valid annotations, totalling roughly 5000 human judgements.

\subsection{From Labels to Distributions}

We fit Beta($\alpha, \beta$) distributions to the annotation intervals using maximum likelihood estimation. We treat the provided annotator intervals $[l_i, u_i]$ as bounding the central probability mass. For each expression, we compute the probability mass $p_i = \text{CDF}_{\text{Beta}}(u_i) - \text{CDF}_{\text{Beta}}(l_i)$ and maximize the log-likelihood $\sum \log(p_i)$. Initial parameters are derived from interval midpoint and width moments and refined via Nelder--Mead optimisation \cite{nelder1965simplex}. The output includes $\alpha$ and $\beta$ parameters, distribution mean and variance, and raw annotation intervals.
Figure \ref{fig:dist-ex} shows examples of the distributions obtained.

%\begin{figure*}[t]
%\centering
%        \includegraphics[width=1\linewidth]{en-distr.png}
%        \caption{Enter Caption}
%\label{fig:dist-ex}
%\end{figure*}

\subsection{Descriptive Statistics}
\begin{figure*}[t]
    \centering
    % First Image: Cropped 50% from the left
    \begin{minipage}{0.48\linewidth}
        \centering
        \includegraphics[scale=0.78,width=\linewidth]{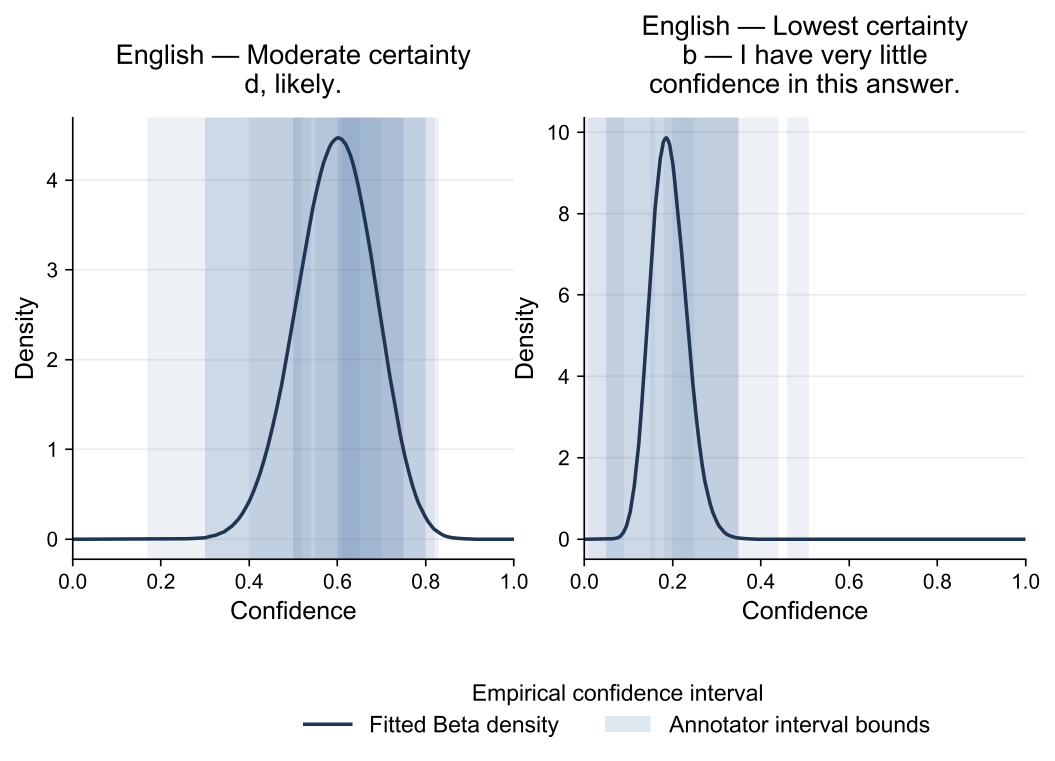}
        \caption{Examples of fitted Beta distributions for English linguistic hedges. The shaded regions represent the empirical interval bounds provided by human annotators.}
        \label{fig:dist-ex}
    \end{minipage}
    \hfill % Adds spacing between the two images
    % Second Image: Normal
    \begin{minipage}{0.48\linewidth}
        \centering
        \includegraphics[scale=0.8, width=\linewidth]{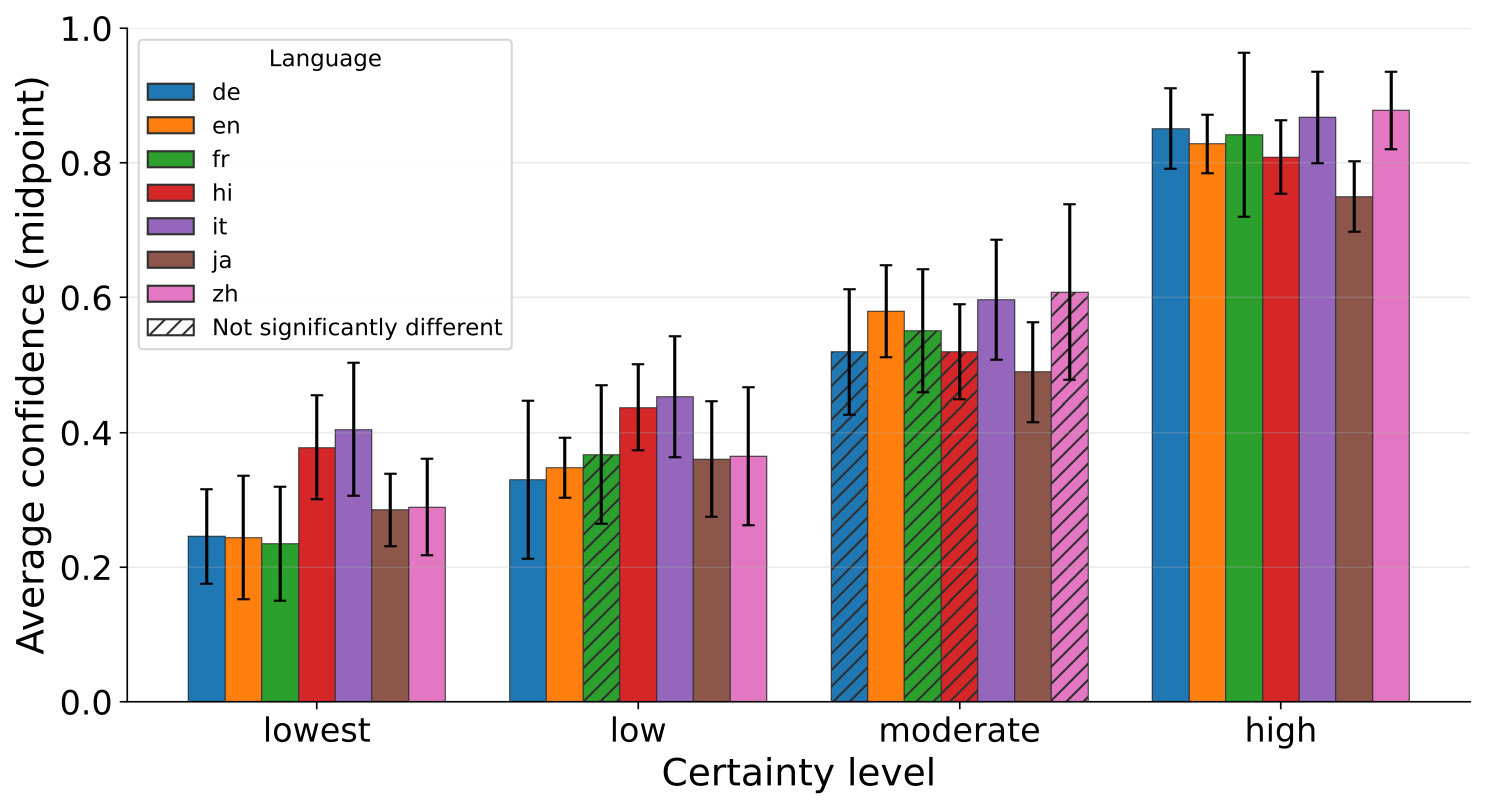}
        \caption{Annotated confidence across German, English, French, Hindi, Italian, Japanese, and Chinese, categorized by four certainty levels. The etched bars do not show statistically significant difference to other languages at the given certainty level. Error bars show +/- 1 std.}
        \label{fig:confidence-stat}
    \end{minipage}
\end{figure*}

Table \ref{tab:language-stats} presents annotation statistics across languages. %Mean interval widths range from 0.216 (Italian) to 0.290 (Hindi), with fluency ratings from 3.981 (French) to 4.435 (Italian) out of 5 and readability from 3.933 (French) to 4.484 (German) out of 5. 
%As Figure \ref{fig:ai-attitude} shows, there seems to be a small, not statistically significant positive correlation between a positive AI attitude and higher average confidence scores being labelled, but a larger annotator pool would provide clearer results.
%Additionally, there seems to be a small, not statistically significant positive correlation between a positive AI attitude and higher average confidence scores being labelled.
Cross-linguistic analysis using Kruskal-Wallis tests \cite{kranich2011hedge} revealed significant differences in expressions' confidence ratings across all certainty levels (Figure~\ref{fig:confidence-stat}). %We selected the Kruskal-Wallis test due to its robustness to non-normal distributions and unequal group sizes across languages, avoiding parametric assumptions. 
Effect sizes were largest at extreme certainty levels (lowest: $\varepsilon^2 = 0.391$; high: $\varepsilon^2 = 0.363$) and smallest at moderate certainty ($\varepsilon^2 = 0.135$). Post-hoc pairwise comparisons with Bonferroni correction \cite{dunn1961multiple} controlled family-wise error rate across 21 language pairs per level, identifying systematic patterns: Italian and Hindi speakers assigned higher confidence values to low-certainty expressions, while Japanese speakers assigned lower values to high-certainty expressions, confirming cultural and linguistic differences (Table~\ref{tab:certainty-full}).

%\begin{figure}[h]        \centering
%        \includegraphics[width=1\linewidth]{image.png}
%                \caption{Enter Caption}
%        \label{fig:ai-attitude}
%    \end{figure}

%\begin{figure}[h]
%        \centering
%        \includegraphics[width=1\linewidth]{conf-stat.png}
%        \caption{Enter Caption}
%    \label{fig:confidence-stat}
%\end{figure}

\begin{table}[t]
\centering
\footnotesize
\setlength{\tabcolsep}{4pt} % Adjust horizontal spacing to fit column
\begin{tabular}{lrrrccccccc}
\toprule
\textbf{Lang} & \textbf{Ann.} & \textbf{Lower} & \textbf{Upper} & \textbf{Width} & \textbf{Flu.} & \textbf{Read.} \\
\midrule
de & 10 & 0.365 & 0.608 & 0.243 & 4.338 & 4.484\\
en & 10 & 0.387 & 0.613 & 0.226 & 4.358 & 4.393 \\
fr & 10 & 0.372 & 0.625 & 0.253 & 3.981 & 3.933 \\
hi & 7 & 0.390 & 0.680 & 0.290 & 4.194 & 4.361 \\
it & 10 & 0.472 & 0.688 & 0.216 & 4.435 & 4.464 \\
ja & 9 & 0.343 & 0.599 & 0.257 & 3.986 & 4.107 \\
zh & 8 & 0.413 & 0.656 & 0.243 & 4.159 & 4.139 \\
\bottomrule
\end{tabular}
\caption{Descriptive statistics for annotations across languages. All mean values are rounded to three decimal places.}
\label{tab:language-stats}
\end{table}

\begin{table}[t]
\centering
\footnotesize
\begin{tabular}{lccc>{\raggedright\arraybackslash}p{0.25\linewidth}}
\toprule
\textbf{Level} & \textbf{$p$-value} & \textbf{$\varepsilon^2$} & \textbf{N} & \textbf{Key Differences} \\
\midrule
Lowest   & $< 10^{-10}$ & 0.391 & 10 & it/hi $>$ de/en/fr; it/hi $>$ ja/zh \\
Low      & $< 10^{-5}$  & 0.207 & 8  & en/de $>$ hi/it; hi/it $>$ ja/zh \\
Moderate & $< 0.001$    & 0.135 & 2  & en/it $>$ ja \\
High     & $< 10^{-9}$  & 0.363 & 8  & de/en/fr/it/hi $>$ ja; zh $>$ ja \\
\bottomrule
\end{tabular}
\caption{Cross-linguistic analysis of certainty expression. Kruskal-Wallis tests with Bonferroni-corrected post-hoc comparisons. N = number of significant language pairs at $\alpha = 0.05$.}
\label{tab:certainty-full}
\end{table}

\section{Training a Confidence Mapper}

To scale the benchmark beyond the annotated set, we train a lightweight multilingual confidence mapper, to go from a natural language expression of confidence to continuous uncertainty representations, parameterised as ($\mu$, $\sigma$) pairs. 
Prior studies leveraged LLMs to estimate the decisiveness of sentences \cite{yona2024largelanguagemodelsfaithfully} or to classify the level of uncertainty \cite{rathi2025humansoverrelyoverconfidentlanguage}. \citet{tao2025largelanguagemodelsexpress} instead developed a lightweight model, using DistilRoBERTa \cite{sanh2019distilbert}, to map expressions to confidence levels.
In this work, we compare three mapping approaches: similarity-based retrieval, a fine-tuned encoder model, and LLM inference.

\subsection{Similarity Mapper}

This non-parametric method stores annotated examples as embeddings and retrieves the k-nearest neighbours for each test expression.
We use multilingual-e5-large-instruct \cite{wang2024multilingual} to retrieve the $k$-nearest training examples for each test expression. Predictions aggregate neighbour targets ($\mu_j, \sigma_j$) via weighted mean or median $\mathcal{A}$:
$$ \hat{\mu}=\mathcal{A}_\mu(\{(\mu_j,w_j)\}), \quad \hat{\sigma}=\mathcal{A}_\sigma(\{(\sigma_j,w_j)\}). $$
Weights are derived from temperature-scaled cosine similarities $s_j$:

$w_j=\exp(s_j/\tau) / \sum_{\ell\in \mathcal{N}_k}\exp(s_\ell/\tau)$.

\subsection{Encoder Mapper}

To adapt \citet{tao2025largelanguagemodelsexpress}'s approach to our multilingual and distributional  setting, we fine-tune xlm-roberta-base \cite{conneau2020unsupervised} with a regression head that maps each model response’s uncertainty phrasing directly to two continuous targets: confidence mean and confidence spread (std).

\subsection{LLM Mapper}

The finetuned LLM mapper is built as an instruction-following causal LM, Llama-3.1-8B-Instruct \cite{grattafiori2024llama}, that is SFT-trained with LoRA \cite{hu2022lora} to translate an uncertainty expression into \{mean, std\}. %During data preparation, each input sentence is masked for answer letters, then paired with continuous targets derived from the human annotations, creating prompt-completion training records. Inference runs the same instruction prompt.

\subsection{Mapper Results}

We evaluate all confidence mappers using 4-fold cross-validation with 3 random seeds on the annotated expressions. Performance is measured using the mean absolute error (MAE) and root mean squared error (RMSE) for both mean and standard deviation predictions, as RMSE heavily penalizes larger errors. %{\color{red}Additionally, we validate the best-performing mapper on the 17 English expressions in \citet{illinoisPerceptionProbability}}.
As Table \ref{tab:model_comparison} shows, the similarity mapper consistently outperforms encoder and LLM-based approaches, with particularly strong performance on mean prediction. 
We hypothesise that, given the relatively bounded lexical nature of linguistic hedges and the limited size of the labelled dataset, non-parametric similarity retrieval performs better than higher-capacity models.

We evaluate calibration metrics for the Linguistic Confidence method using the gold human-annotated distributions on the held-out data from each fold. Using gold distributions yields an Expected Calibration Error of 0.49, compared to 0.48 when using the similarity mapper. 
We report hyperparameter tuning and per-language results in Appendix \ref{sec:mappers}. We adopt the best-performing mapper for all subsequent uncertainty calibration experiments. 

\begin{table}[htbp]
\centering
\resizebox{0.4\textwidth}{!}{% % Adjust 0.8 to 1.0 for full page width, or down to 0.6 to make it smaller
\begin{tabular}{lcccc}
\toprule
\textbf{Model} & \textbf{MAE $\mu$} & \textbf{MAE $\sigma$} & \textbf{RMSE $\mu$} & \textbf{RMSE $\sigma$} \\
\midrule
Similarity & \textbf{0.075}& \textbf{0.039}& \textbf{0.101}& \textbf{0.051}\\
Encoder & 0.152
& 0.050
& 0.189
& 0.062
\\
LLM & 0.183
& 0.067
& 0.244
& 0.071
\\
\bottomrule
\end{tabular}%
}
\caption{Comparison of mapper models. Best in bold.}
\label{tab:model_comparison}
\end{table}

\section{Linguistic Confidence in LLMs}

Recent research \cite{tao2025revisiting} shows that linguistic confidence is often better calibrated than other black-box single-pass methods, while consistency-based methods like multisampling are calibrated but not accessible \cite{wang2022self}.
We compare four distinct signals: models' logits, answer frequency in multisampling, the model's verbalized numerical confidence, and the produced linguistic hedge with the associated mapped distribution. 
We sample the same 1,000 questions per language from GlobalMMLU Lite \cite{singh2025global}, ensuring no overlap with our labelled data. %We validate our results on open questions from WebFAQ 2.0 \cite{dinzinger2026webfaq}, a multilingual FAQ-based natural question-answer dataset. These results and the prompts used for inference are reported in Appendices \ref{sec:results} and \ref{sec:prompts}.
The prompts used for inference are reported in Appendix 
\ref{sec:prompts}.
Additionally, we validate our results for token-probability and linguistic confidence in a free-form generation setting for instruct models, presenting the questions without the answer options.%; these results are detailed in Appendix \ref{sec:openqa}.

\subsection{Confidence Elicitation Methods}

\begin{figure*}[h!]
    \centering
    \includegraphics[width=1\linewidth]{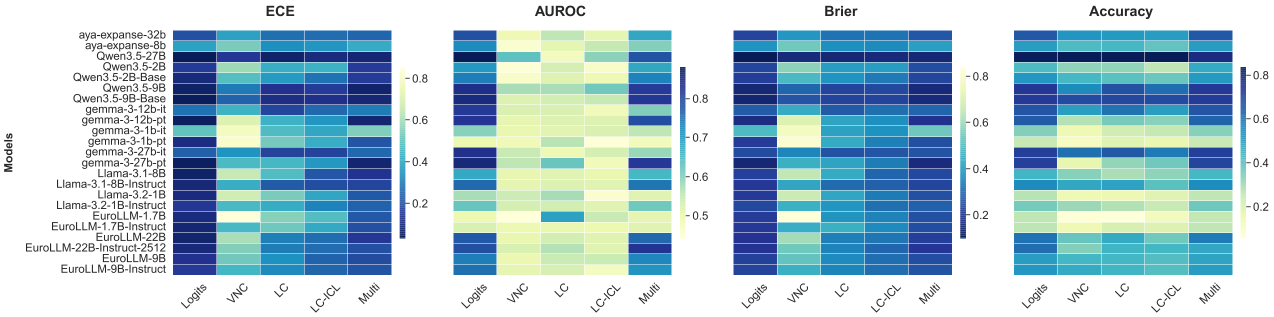}
    \caption{Cross-model evaluation of confidence elicitation methods across different uncertainty metrics on MCQ data. Each heatmap evaluates internal logits (logit), verbalised numerical confidence (VNC), linguistic confidence in zero shot (LC) and 5-shot (LC-ICL) settings, and multisampling (multi). Performance is measured using Expected Calibration Error (ECE), Area under the ROC Curve (AUROC), Brier Score, and task Accuracy.}
    \label{fig:heatmaps}
\end{figure*}

%Language models may exhibit different calibration when expressing probabilities directly, through generated text, or through repeated sampling. To capture these varied signals, we employ four distinct confidence elicitation strategies:

\textbf{Logit-based Confidence.} Accessing the model's internal next-token probabilities provides a direct, low-level measure of its baseline certainty for MCQ data. Let $z$ be the model's next-token logits at the answer slot. We compute the vocabulary log-probabilities $\ell_t=\log\mathrm{softmax}(z)_t$, and normalize over the valid options $k\in\{A,B,C,D\}$ to extract the predicted answer $\hat y$ and confidence $c_{\text{logit}}$:
$$p_k=\frac{e^{\ell_k}}{\sum_{j\in\{A,B,C,D\}} e^{\ell_j}}.$$
From this, we extract the predicted answer $\hat y=\arg\max_k p_k$ and the confidence $c_{\text{logit}}=\max_k p_k$.

When running validation experiments on free-form generation, as we do not have option answers, we look at the mean probability of generated answer tokens \cite{geng2024survey, vazhentsev2023hybrid}. 

\textbf{Multisampling (Consistency).} We approximate confidence through the variance of the model's responses. We sample the model $N=10$ times for the same question. If $n_k$ is the frequency count of option $k$ across all samples, the prediction and confidence are defined by majority agreement:
$$\hat y=\arg\max_k n_k,\qquad c_{\text{multi}}=\frac{\max_k n_k}{N}.$$

\textbf{Linguistic Calibration (LC).} Models are pre-trained on vast amounts of human text, where uncertainty is naturally expressed through qualitative hedging rather than strict numbers. We employ both zero-shot and 5-shot ICL variants to prompt the model for hedged, confidence-aware answers. During evaluation, the LC expression is mapped to a beta distribution via the similarity mapper.

\textbf{Verbalized Numerical Confidence (VNC).} We prompt the model to generate its answer alongside a verbalized percentage confidence (e.g., ``$73\%$"). This tests whether the model can accurately articulate its certainty without requiring access to its internal weights.

\subsection{Evaluation Metrics}

A well-performing model must not only be accurate but also well-calibrated \cite{guo2017calibrationmodernneuralnetworks} and highly discriminative \cite{hendrycks2016baseline}. Calibration measures how well predicted confidence aligns with answer correctness, and discriminability assesses whether a model can recognise correct from incorrect answers. In addition to accuracy, we report:

\textbf{Expected Calibration Error (ECE).} For a set of confidence-correctness pairs, we partition the confidence space $[0,1]$ into $B$ equal-width bins and compute the weighted average of the absolute difference between empirical accuracy and predicted confidence for each bin $b$, where $I$ is the count of predictions in that bin:
$$\mathrm{ECE}=\sum_{b=1}^B \frac{|I_b|}{N}\,\bigl|\mathrm{acc}(b)-\mathrm{conf}(b)\bigr|.$$

ECE treats confidence as a point mass, which fails to capture the shape and spread of predictive uncertainty. To account for this, we utilize an adapted approach \cite{wang2025calibratingexpressionscertainty} for methods that produce confidence as Beta distributions.
For each sample $n$ with predicted confidence distribution $g(x_n) \sim \text{Beta}(\alpha_n, \beta_n)$, it contributes to all bins with weights determined by $P(S \in I_m \mid X_n)$, the probability that a random draw $S$ falls within the interval $I_m$, given the specific distribution produced for sample $X_n$.
As \citet{wang2025calibratingexpressionscertainty} show, traditional ECE is a restricted special case of this framework, simplifying the generalised version directly back into the single-bin summation of ECE.
%For each bin, we compute two key quantities: (1) the average accuracy of samples contributing to that bin, and (2) the average expected confidence within that bin. The accuracy contribution is a weighted sum, where each sample is weighted by the probability mass of its confidence distribution falling within the bin. The confidence contribution is computed by directly integrating the expected value of the sample's distribution over the specific interval of the bin. The distributional ECE is then the weighted sum of absolute differences between the normalized confidence and accuracy across all bins, where $m_b$ is the probability mass in bin $b$:
%$$\mathrm{DistECE}=\sum_b w_b\,\bigl|\overline{c}_b-\overline{a}_b\bigr|,\qquad w_b=\frac{m_b}{\sum_{b'} m_{b'}}.$$
%This estimator has been shown to be consistent and more robust to bin granularity than classical binned ECE, as the soft binning mechanism ensures all samples contribute information to every bin rather than being hard-assigned to a single bin \citep{wang2025calibratingexpressionscertainty}.

\textbf{Area Under the ROC Curve
(AUROC) for discriminability.} We compute the AUROC to determine model discriminability. Using confidence as the score and correctness as the binary label, AUROC represents the probability that a correct sample receives a higher confidence score than an incorrect one.
\textbf{Brier Score.} Additionally, we report the Brier Score, a proper scoring rule which measures the mean squared probabilistic error. %As a strictly proper scoring rule, a lower Brier Score indicates superior performance, simultaneously penalizing poor calibration and a lack of predictive sharpness.

\subsection{Model selection}
We evaluate a range of Transformer-based open-source multilingual LLMs, comprising 23 variants from five model families, in the base and instruct variants, ranging from 1 to 32 billion parameters: Qwen3.5 \cite{bai2023qwen}, Aya Expanse \cite{dang2024aya}, Llama 3 \cite{grattafiori2024llama}, Gemma 3 \cite{team2024gemma}, and EuroLLM \cite{ramos2026eurollm}.
We run inference using models' default parameters on NVIDIA-A100 GPUs.

\subsection{Results}

%The primary objective of our experiments was to identify discrepancies between a model's internal probability states and its externalized linguistic outputs.

\subsubsection{Comparing Elicitation Methods}

We find that in the MCQ setting the logit-based method is the most effective for uncertainty quantification. This method achieves superior results in calibration and discriminability across models (Figure \ref{fig:heatmaps}).  As the validation on 8 to 12B instruct models detailed in Appendix \ref{sec:openqa} shows, though, logit-based methods are not as well-calibrated in free-form generation, as they represent a signal of text fluency rather than a confidence probe, while linguistic confidence achieves better performance (0.537
ECE for mean token probability vc. 0.490 for LC). 
In aggregate results, multisampling is the second most competitive strategy, but Linguistic Confidence (LC) and the in-context learning variant (LC-ICL) show very competitive calibration performance in specific models.
Overall, calibration is misaligned as models attempt to "surface" their uncertainty into verbalised formats.
%This phenomenon is evidenced by the poor calibration of VNC, followed by LC, in comparison to logits in the MCQ setting. 
%In Appendix  \ref{sec:results} we analyse per-language performance for LC and LC-ICL.

\begin{figure}[h!]

    \begin{minipage}{0.45\textwidth}
%\end{figure}
%\begin{figure}
        \centering
 \includegraphics[width=0.85\linewidth]{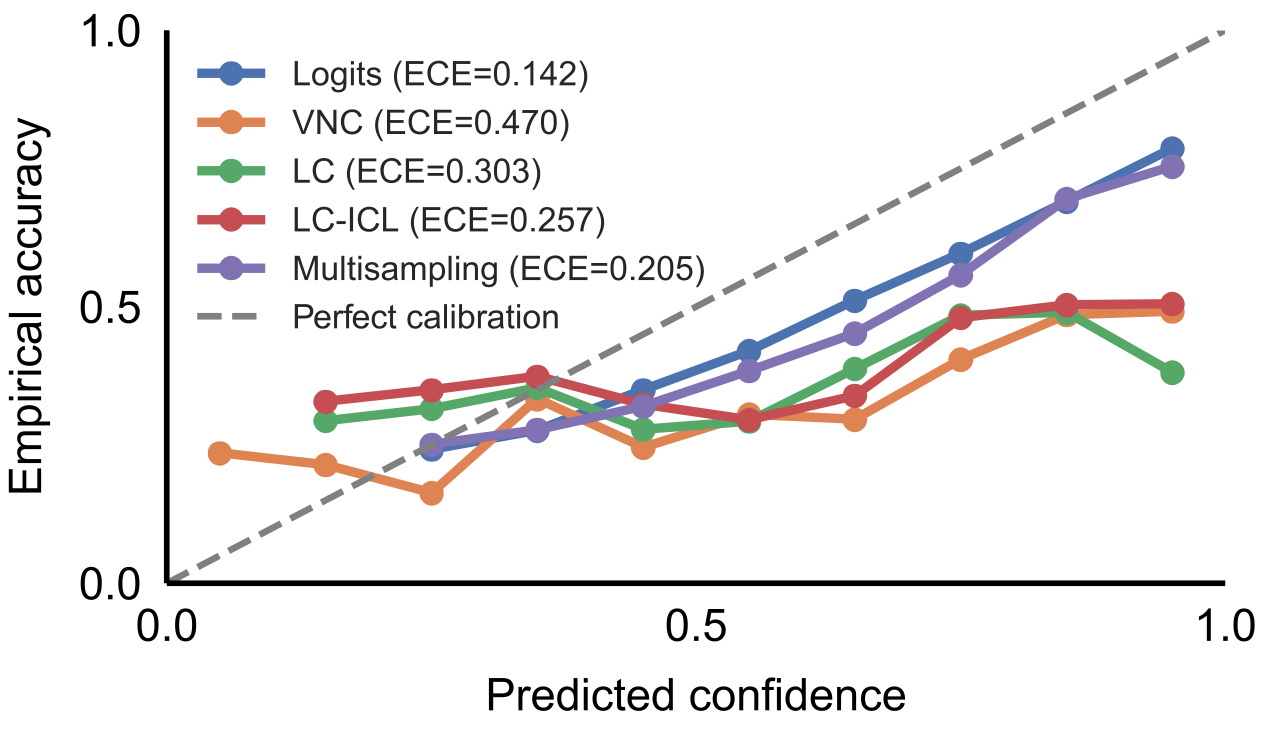}
        \caption{Calibration plot comparing empirical accuracy against predicted confidence across models. }
        \label{fig:cal-curve}
        \end{minipage}    \label{fig:placeholder}
\end{figure}

Additionally, % as Figure \ref{fig:abcd-comp} shows, 
 logit-based ECE is better even when comparing the logits for the option letter under the base prompt to those generated together with a linguistic expression of confidence (mean ECE for 0.11 for base logits vs. 0.28 for logits in LC-ICL).
The calibration plot in Figure \ref{fig:cal-curve} highlights that the verbalised methods can not only be overconfident, but also underconfident, with a confidence lower than 0.25 for questions with only four possible answer options. Additionally, VNC is the worst calibrated method, and  models seem to favour scores multiple of 5 (eg. 25\%, 50\%, 75\%). 
%Worse calibration with verbalised methods like LC and VNC is an issue, as these methods are available for commercial models, more user-friendly and cost-effective, but the miscalibrated and overconfident responses generated can misplace user trust.

\begin{figure*}[h!]
    \centering
    \includegraphics[width=1\linewidth]{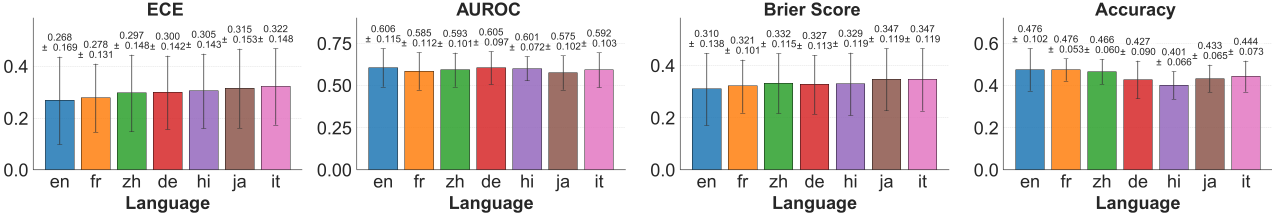}
    \caption{Per-language ECE, AUROC, Brier, and Accuracy averaged across methods and models, +/- 1 std.}
        \label{fig:language-comp}
\end{figure*}

\subsubsection{Cross-Linguistic Patterns}

Figure \ref{fig:language-comp} shows that English performs best across all metrics. %Conversely, languages like Hindi, German, and Italian often struggle, either in terms of raw accuracy or calibration.
A pairwise statistical analysis (Appendix \ref{sec:stat}) shows that, with respect to ECE, Brier, and accuracy, English performs significantly better than all other languages apart from French, where the difference is not significant. With respect to AUROC, fewer pairs are significant, and English performs significantly better than French (q=0.002), Italian (q=0.028), Japanese (q=4.16e-07), and Chinese (q=0.011).
We find that sensitivity to language variation depends on the elicitation method and that, while the performance gap between English and other languages is present across model families, we do not identify any significant differences.
Additionally, through a fixed-regression analysis (Appendix \ref{sec:stat}), we confirm that language remains associated with calibration even after controlling for task accuracy.
%For Hindi, accuracy is the lowest, while the calibration is worst for German, Hindi, and Italian, suggesting that for these languages the model's predicted probabilities are less representative of the actual likelihood of correctness.
%Brier score closely follows the same patterns as ECE.
%When looking at discriminability, AUROC shows narrower gaps between languages: English leads, closely followed by German.
%In Appendix \ref{sec:stat} we show that, while the logit-based and multisampling methods show narrower gaps, the verbalised methods are subject to higher sensitivity to the language used. 

\subsubsection{Instruction Tuning and Model Size}

\begin{figure}[!h]
        \begin{minipage}{0.48\textwidth}
        \centering
               
                    \centering
                    \includegraphics[width=0.9\linewidth]{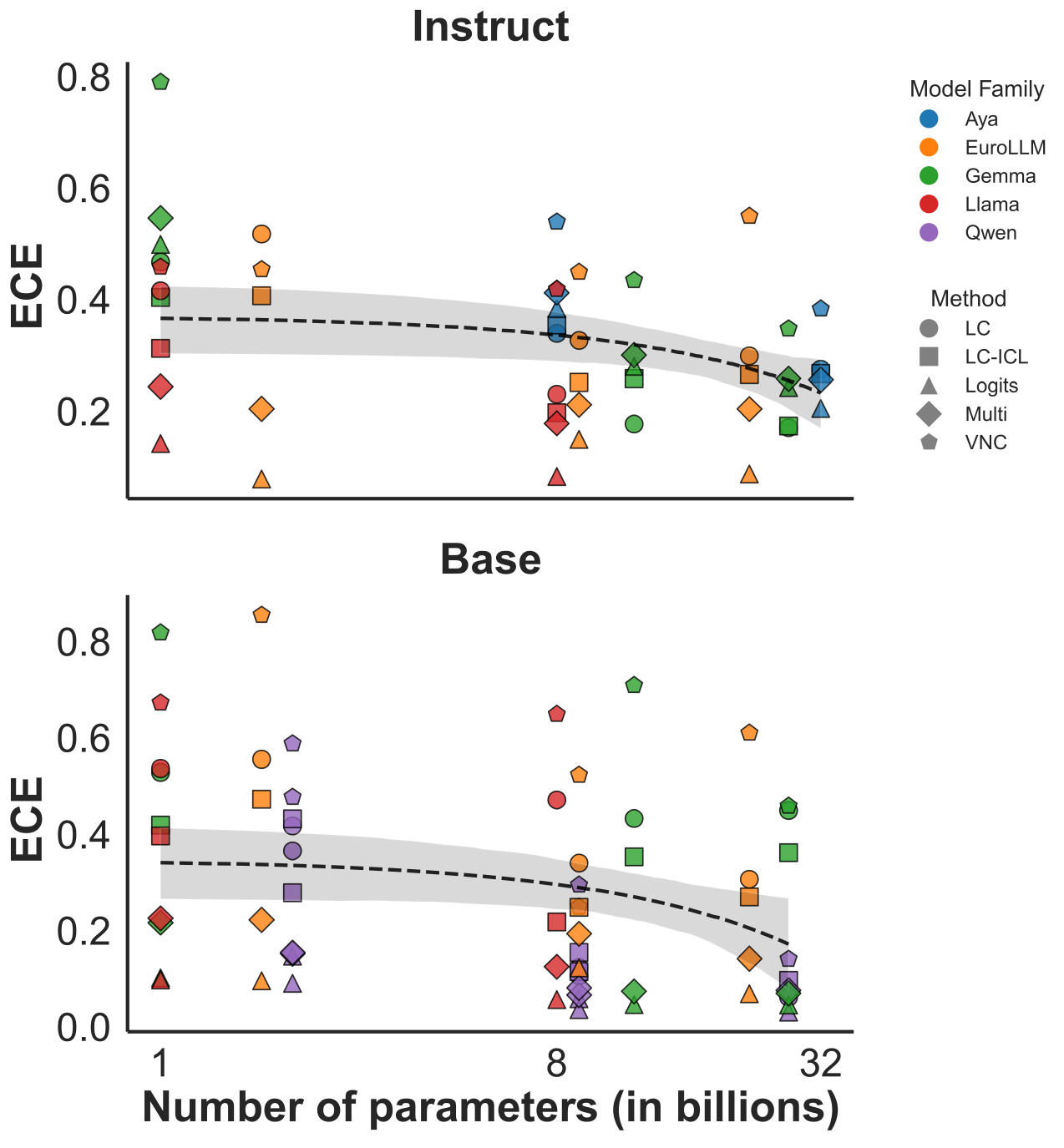}
                    \caption{ECE across different model sizes with 95\% CI regression line. Each point represents a model and elicitation method.}
                    \label{fig:size}

    \end{minipage}
    \end{figure}

For internal measures, such as logits and multisampling, instruction tuning generally improves discriminability, but simultaneously worsens calibration. In contrast, for methods relying on verbalised confidence, instruction tuning overall aids calibration, due to the requirement for instruction-following and structured output.
%The most dramatic gains were observed in the VNC approach, where instruction tuning led to substantial improvements in ECE and Brier Score. Similarly, for LC using ICL, post-training reduced ECE and Brier Scores while improving AUROC. 
Additionally, Figure \ref{fig:size} shows improvement in calibration with an increase in model size, across elicitation methods and instruction-tuning.

%\begin{figure}[h!]
  %  \centering
  %  \includegraphics[width=0.8\linewidth]{cal-curve.png}
  %  \caption{Calibration plot comparing empirical accuracy against predicted confidence, averaged across instruction-tuned models.}
 %        \label{fig:cal-curve}
%\end{figure}

%\begin{figure}[htbp]
%     \centering
%     \begin{minipage}[b]{0.47\linewidth}
%         \centering
%         \includegraphics[width=\linewidth]{cal-curve.png}
%         \caption{Calibration plot comparing empirical accuracy against predicted confidence, averaged across instruction-tuned models.}
%         \label{fig:cal-curve}
%     \end{minipage}
%     \hfill
%     \begin{minipage}[b]{0.47\linewidth}
%         \centering
%         \includegraphics[width=\linewidth]{abcd_comp.png}
%         \caption{Mean Expected Calibration Error (ECE) comparing base logit extraction against the logit-based uncertainty for the answer letter in LC.}
%         \label{fig:abcd-comp}
%     \end{minipage}
%     \
%     \label{fig:combined_metrics}
%\end{figure}

\section{Discussion}
% do models get better calibrated as they grow in parameter size?
% are newer models better calibrated?
% are languages performing better if closer to how english uses hedges? this seems not to be the case, but there are many directions it could looked at
Internal, verbalised, and perceived confidence do not align. %Specifically, as confidence expressions are verbalised, model calibration typically worsens. 
While internal token logits provide the most calibrated measure of uncertainty in the MCQ setting, they are inaccessible to users and not well-suited to free-form generation. Linguistic Confidence is more user-friendly and better calibrated than Verbalised Numerical Confidence, but still requires extensive critical evaluation.
The confidence ranges humans assign to hedging words do not natively match LLMs confidence for the corresponding output. Mapping Linguistic Calibration to probability distributions provides a useful measure of the vagueness and fuzziness of language, but evaluation confirms that models cannot faithfully translate their uncertainty into words.

The performance gap between English and other languages raises questions about the trustworthiness of LLMs in highly multilingual and global contexts and about whether LLMs can use hedges and epistemic markers in different languages without losing important cultural and linguistic differences. 
Because human interpretation is fuzzy, asymmetric, and culturally dependent, mapping linguistic calibration to probability distributions is essential.
The fact that English consistently leads in calibration suggests that %models' understanding of confidence is optimised for English linguistic patterns, and 
languages with very different strengths and prevalences of hedges may be left behind.

\section{Conclusion}

%Well-calibrated linguistic expressions of uncertainty in multilingual LLMs are necessary due to the quickly growing number of users interacting with LLM-based chat interfaces or LLM-augmented search engines on a daily basis, even with high-stakes queries. Currently, however, while models possess relatively well-calibrated internal measures of uncertainty, their surface expressions are often miscalibrated, overconfident, and inconsistent across languages.

This study provides a distributional and multilingual study of linguistic confidence, treating hedges as continuous probability distributions and moving beyond English-centric evaluations. Verbalised and logit-based measures for calibration are persistently misaligned. 
Additionally, instruction tuning may improve the structure of confidence expressions while distorting the model's internal calibration.
The fact that confidence scores from different methods diverge reinforces the need for more reliable, multilingual, calibration methods.
We argue that a distributional framework for linguistic confidence is necessary for ambiguous and vague expressions, and it offers a more robust, human-centric way to interpret uncertainty than numerical confidence. 
Through labelled data, we note cross-linguistic variation is significant, but models achieve worse performance in other languages than in English.
%Our proposed multilingual benchmark and confidence mapper are steps toward ensuring that LLM outputs carry the same perceived uncertainty across different linguistic norms. 
Future work should involve diverse users to better understand the linguistic and cultural nuances of confidence, trust, and uncertainty expression, fostering transparency and safety in multilingual human-AI interactions.

\section*{Limitations}
Our experiments are conducted on small to medium-sized models with MCQ data, and only validated on a sample of open questions for few models.
However, the user-centered value of linguistic hedges is most valuable in open-ended dialogue and ambiguous interactions. Future research should evaluate distributional Linguistic Confidence in free-form generation settings across varied domains.

While our similarity mapper achieves strong performance with minimal error propagation, its reliance on multilingual embeddings could impose an English-centric semantic similarity. This risks distorting cross-lingual hedge interpretation for non-Western languages with vastly different pragmatic structures. 

Additionally, while the distributional labels in our benchmark are derived from Prolific crowdworkers, the initial generation and validation of the multilingual hedges relied heavily on proprietary LLMs. As we did not conduct validation with native speakers for all seven languages prior to annotation, some language-specific nuances or errors in the generated hedges may have been retained.

Finally, despite our dataset being the most comprehensive distributional multilingual hedge collection so far, it is small by the standards required for model training or fine-tuning. The high cost and complexity of gathering multi-annotator belief intervals constrained the dataset's size. Therefore, the full long tail of hedging vocabularies in each language is limited. Future work should prioritize expanding this data with attention to cultural and linguistic validity.

\section*{Ethical Considerations}
We hope this work advances the calibration of linguistic confidence in multilingual large language models, and further motivates the design of fairer multilingual LLMs and of safety evaluations that account for cross-linguistic variation and human experiences of AI generated language.
Nonetheless, the languages we focused on are still considered high-resource, as these were the most easily available for crowdsourcing annotations. Overconfidence risks will still arise for lower-resource languages, where model calibration might be harder to quantify if their linguistic expressions of confidence remain understudied.
Additionally, a significant risk of English-centric bias remains, as LLM-generated text, used for the expansion of the dataset's answers to include linguistic hedges, often follows Anglocentric structures and communication patterns.

In real-world deployment, there is a danger of human overreliance on LLMs. As models frequently provide overconfident responses, the persuasive and fluent use of hedging or expressions of confidence, without proper calibration and pragmatic competence can lead to misplaced user trust. Finally, the expression of confidence and uncertainty through linguistic expressions inherently carries the risk of anthropomorphism. The development of AI systems that use natural linguistic hedges and express uncertainty in a way that mimics human cognition more closely could further blur the line between machine output and human interaction.

\section*{Acknowledgements}
This work was supported by the UKRI AI Centre for Doctoral Training in Designing Responsible Natural Language Processing, grant number EP/Y030656/1. 
The annotation study was approved by the University of Edinburgh School of Informatics Ethics Committee, Application 909945.
The authors used GitHub Copilot to assist with code completion. All code was manually verified and tested by the authors.
The authors would like to sincerely thank Mugdha Pandya (University of Edinburgh) for validating the annotation instructions in Hindi and providing valuable feedback. 

\bibliography{custom}

\appendix

\section{Dataset Augmentation and Annotation Methodology}
\label{sec:annotation}

This appendix provides an overview of the procedures used to augment our dataset with multilingual expressions of varying confidence levels, the quality control applied, and the design of the human annotation survey.

The dataset we build on, GlobalMMLU Lite \cite{singh2025global}, is distributed under an Apache 2.0 license. Our use of the dataset to evaluate multilingual uncertainty is fully consistent with its intended use.
The human-annotated multilingual hedge dataset produced in this work will be made publicly available under a CC-BY 4.0 license upon publication.

\subsection{Dataset Augmentation}

To study how linguistic confidence is expressed across different languages, we prompted GPT-5 mini to generate answers to multiple-choice questions reflecting varying degrees of certainty. This augmentation is performed across seven languages: English (en), French (fr), German (de), Japanese (ja), Mandarin (zh), Hindi (hi), and Italian (it).

Following \citet{tao2025largelanguagemodelsexpress}, for each ground-truth answer, the model is instructed to generate 50 distinct answer sentences distributed evenly across five target confidence levels.
The prompts explicitly instructed the model to vary its wording across the different levels, focus on natural and diverse expressions, and output the correct answer choice letter (a, b, c, or d) rather than the literal content of the option.

\begin{tcolorbox}[colback=gray!5!white,colframe=gray!75!black,title=\textbf{Data Augmentation Prompt \\(Engish)}, arc=2mm, left=2mm, right=2mm, top=2mm, bottom=2mm]
You are given a multiple-choice question and its ground-truth answer. Your task is to generate 50 answer sentences that express the same answer using different levels of confidence, in English: \\
10 with high confidence, \\
10 with moderate confidence, \\
10 with low confidence, \\
10 with lowest confidence, \\
10 with complete uncertainty, reject to reply. \\
The wording should vary across the levels. Focus on natural and diverse expressions of confidence. You should answer using the letter of the correct answer choice (a, b, c, or d) rather than the content of the answer choice itself.

\end{tcolorbox}

\subsection{Data Quality and Refinement}

To ensure the generated text resembled natural, conversational human-AI interactions rather than templated text, the sampled data in English and Italian was validated manually. All other languages were validated through backtranslation, and an automated LLM-based data refinement step was introduced. 

\begin{tcolorbox}[colback=gray!5!white,colframe=gray!75!black,title=\textbf{Example Data Refinement Prompt (Italian)}, arc=2mm, left=2mm, right=2mm, top=2mm, bottom=2mm]
\small
Improve all rows according to the instructions.\\
Return ONLY strict JSON.\\
Output format: \texttt{\{"results":[\{"item\_id":"...",\\"improved\_text":"..."\}]\}}\\
Include exactly one result per input row, preserving item\_id.\\
Each improved\_text must be a single cleaned response text.

\vspace{0.5em}
\textbf{Instruction:}\\
Refine these Italian responses to sound more like a natural conversation.

Fix literalisms: Change awkward phrasing like `b come opzione' to more natural versions like `indicherei la b' or `opterei per b'.

\vspace{0.5em}
\textbf{Global constraints:}\\
Correct any mistakes in the use of the language. If the answer is already fluent and natural, do not change it. Keep each rewritten response brief (about one short sentence). Preserve the original certainty level and intended answer letter. Avoid adding extra reasoning or chain-of-thought. Across the batch, maintain lexical diversity: do not reuse the same opening template repeatedly.

\vspace{0.5em}
\textbf{Rows:}\\
\texttt{[\{"item\_id": "it\_batch\_01\_row\_01", "text": "Sono molto sicuro e scelgo b come opzione."\}, \{"item\_id": "it\_batch\_01\_row\_02", "text": "Ho una bassa certezza ma la risposta è c."\}]}
\end{tcolorbox}

\textbf{Language-Specific Instructions} 
In addition to the shared constraints, specific instructions were provided to correct common artifacts generated by the initial augmentation phase:
\begin{itemize}
    \item \textbf{French (fr) \& Hindi (hi):} Instructed to remove parenthetical or slashed gender markers (e.g., \textit{sur(e)}, \textit{karunga/karungi}) and standardize to a single form.
    \item \textbf{Italian (it):} Focused on removing literalisms and awkward phrasing. For example, converting "b come opzione" to the more natural "indicherei la b" or "opterei per b".
    \item \textbf{Japanese (ja):} Instructed to avoid overly stiff or cold refusals like "\ja{回答を拒否します}" (Refuse to answer).
    \item \textbf{German (de) \& Mandarin (zh):} Focused on removing unnatural technical terminology. In German, terms like "Unsicher gewichtet" were replaced with "Ich bin mir unsicher". In Mandarin, templated phrasing like "\zh{答案标记为} c" were replaced with natural conversational phrases.
\end{itemize}

\subsection{Human Annotation Interface and Task Design}

\begin{figure*}
    \centering
    \includegraphics[width=1\linewidth]{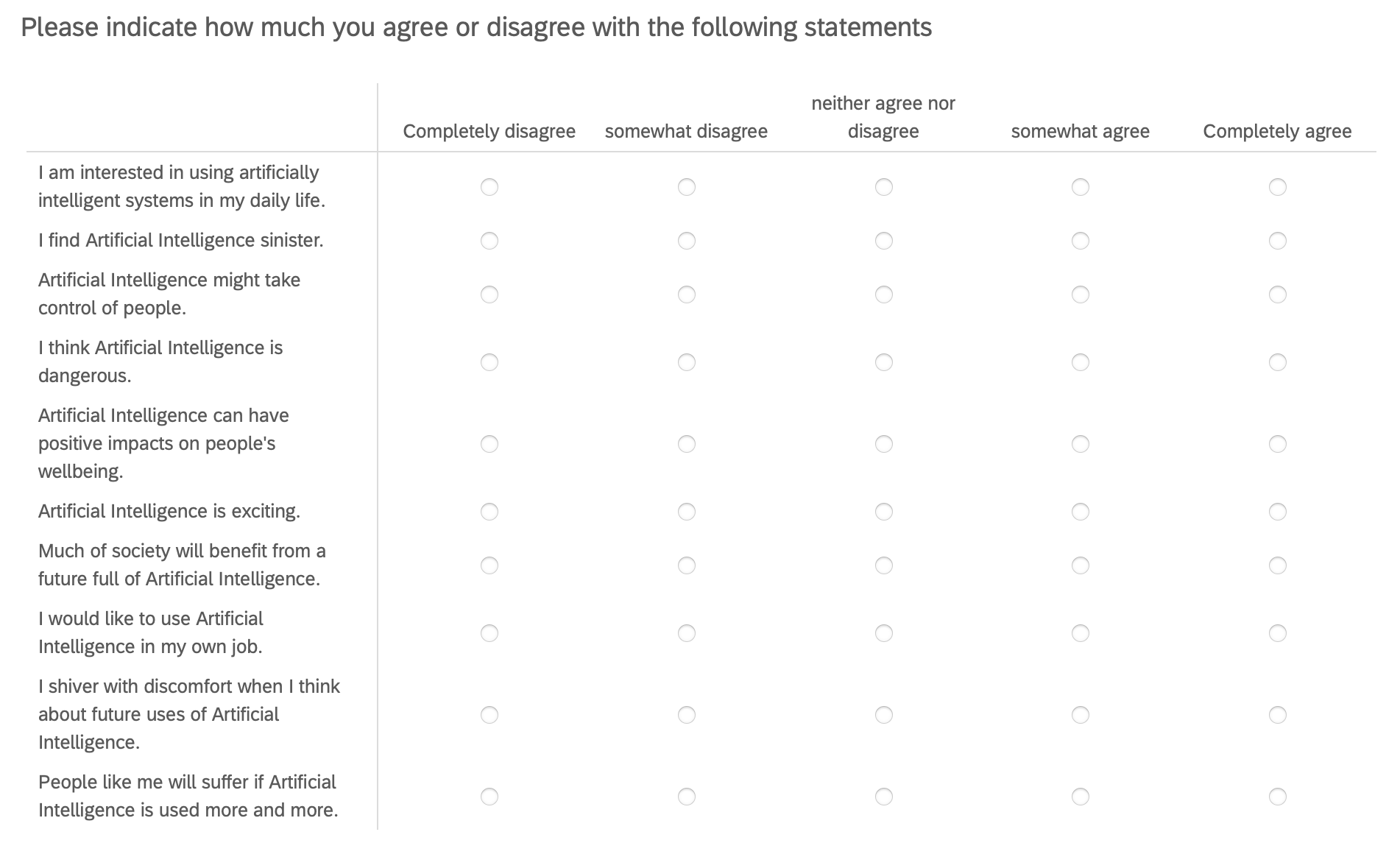}
    \caption{The 10-item questionnaire used to assess participants' general attitudes and biases toward Artificial Intelligence.}
    \label{fig:attitude}
\end{figure*}

To evaluate how users perceive the confidence levels expressed in the generated text, we designed a comprehensive annotation survey through Prolific. 
The study was approved by the ethics board at the authors' institution, and participants filled in an informed consent and participant information sheet when agreeing to take part in the study.
We recruited 10 annotators per language,  self-reporting as native speakers, with a working proficiency of English. Each annotator received £10 for an estimated survey duration of 50 minutes, for a total of £980.
The survey was structured to capture both the perceived probability of correctness and the overall linguistic quality of the response. Additionally, we collected the annotators' general attitudes toward AI, using \citet{Schepman06012026}'s validated questionnaire (Figure \ref{fig:attitude}).

\begin{figure*}[!htbp]
    \centering
    
    % First Survey Image: Instructions
    \begin{minipage}{\linewidth}
        \centering
        \includegraphics[width=1\linewidth]{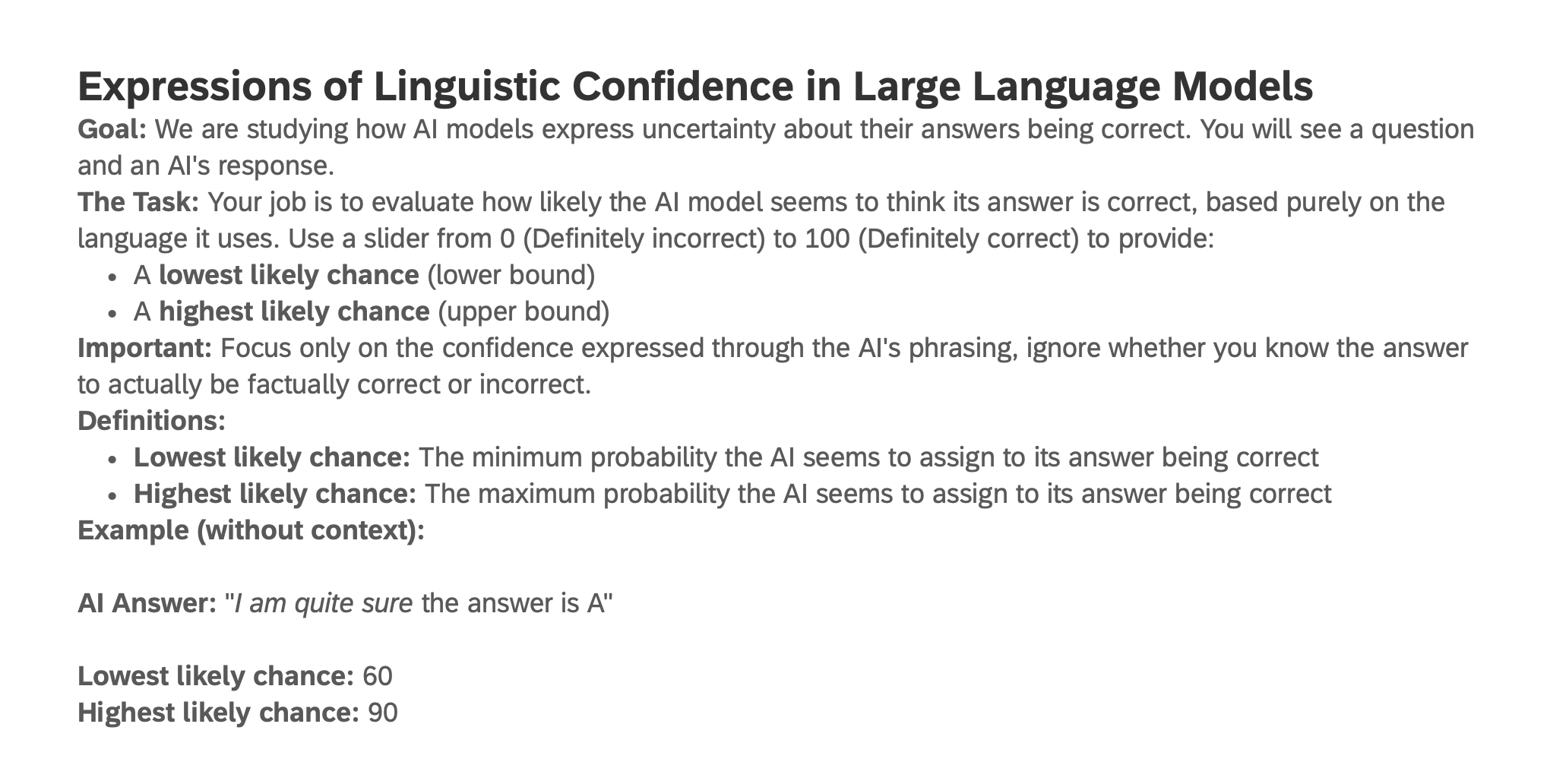}
        \caption{Instructions and definitions provided to annotators for evaluating the linguistic confidence of the AI models (English version).}
        \label{fig:survey_instructions}
    \end{minipage}
    
    \vspace{2em} % Adds vertical space between the images
    
    % Second Survey Image: Sliders and Fluency
    \begin{minipage}{\linewidth}
        \centering
        \includegraphics[trim=5cm 0 0 0, clip, width=0.8\linewidth]{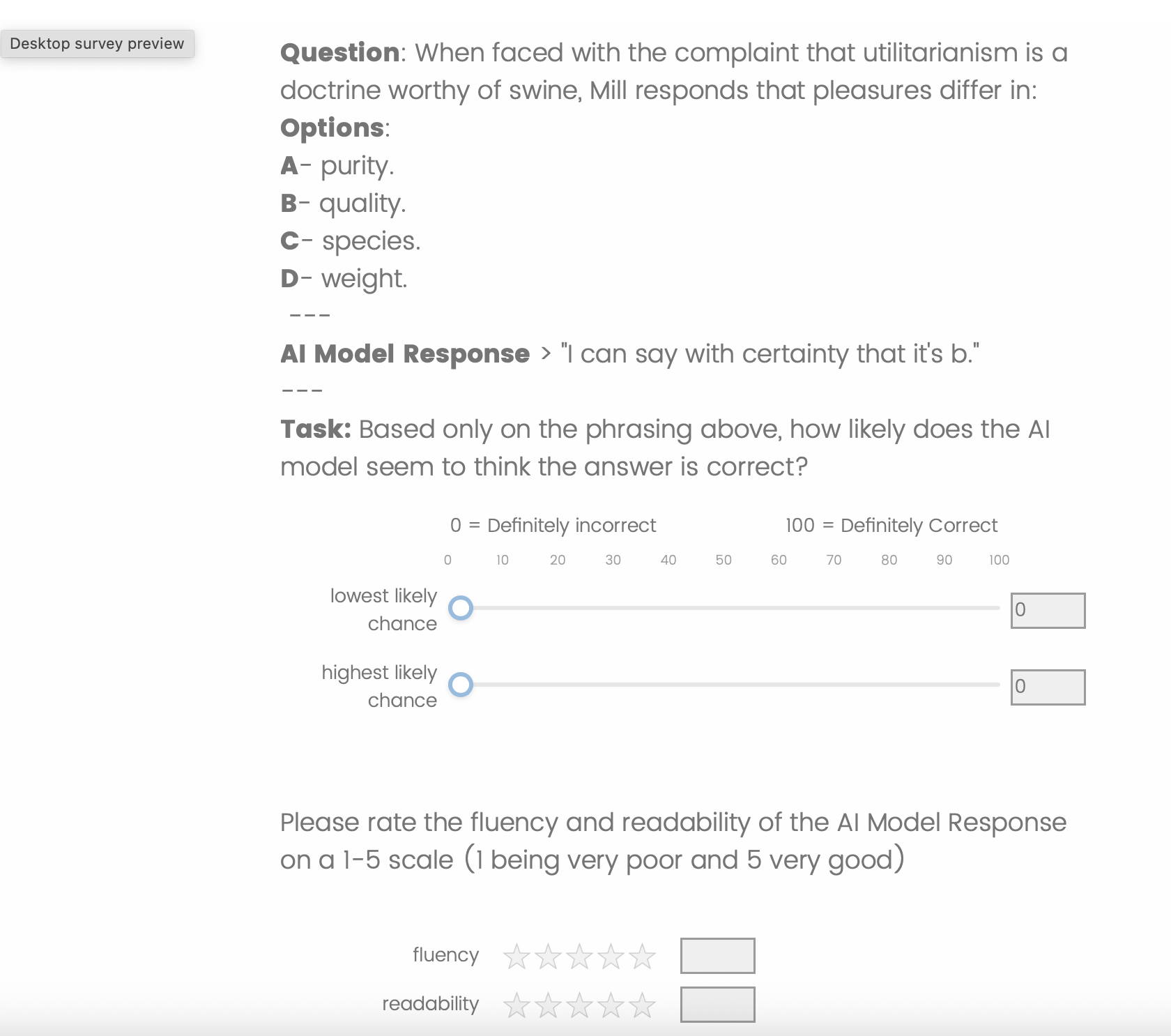}
        \caption{The core annotation interface, featuring continuous sliders for confidence bounds and Likert scales for fluency and readability (English version).}
        \label{fig:survey_interface}
    \end{minipage}
\end{figure*}

Annotators were presented with a multiple-choice question, its options, and the augmented AI response. They were instructed to evaluate how likely the AI model seemed to think its answer was correct, based only on the phrasing. Annotators provided their assessment using two continuous sliders ranging from 0 (Definitely incorrect) to 100 (Definitely correct):
\begin{itemize}
    \item Lowest likely chance: The minimum probability the AI seemed to assign to its answer.
    \item Highest likely chance: The maximum probability the AI seemed to assign to its answer.
\end{itemize}
Annotators were then asked to rate the linguistic quality of the AI's response on a 1--5 scale (represented by stars, where 1 is very poor and 5 is very good) across two dimensions:
\begin{itemize}
    \item Fluency: How natural the text sounded.
    \item Readability: How easy the text was to comprehend.
\end{itemize}
The interface is shown in Figures \ref{fig:survey_instructions}, \ref{fig:survey_interface}.

\subsection{Labelled data validation}
\label{sec:data_val}
We analyse inter-annotator agreement for the labelled intervals to check for disjoint, bimodal belief clusters. In 88.57\% of cases, annotator intervals overlap to form a contiguous block. Disjoint intervals occur in only 11.43\% of expressions.

For English, 11 expressions are two-component disjoint, and 8/11 are likely outlier-driven (1 minority annotator only). The 3 cases where two or more annotators fall in the minority interval are the following expressions: \emph{D — I barely have any confidence in this.}, \emph{I guess d.}, and \emph{a. There's a chance this is correct.}
Italian shows a similar pattern, where 10/11 are single minority outlier splits.
Japanese, in contrast, shows 8/20 expressions where two or more annotators labelled intervals in the minority split: For example, in \emph{\ja{断言はできませんが、a かもしれません。}} (\emph{I can’t assert it, but it may be a.}), and in \emph{\ja{本当に自信はないのですが、一応Bとします。}} (\emph{I really don’t have confidence, but I’ll go with B for now.}), annotators appear to disagree on whether double hedging should map to near-zero confidence or to low–moderate uncertainty.
This suggests most disjoint cases reflect small annotator calibration variance, while true disagreement is concentrated in a small subset where mixed signals induce differences in interpretation.

\begin{table}[h]
\centering
\begin{tabular}{lc}
\hline
Language & Rate\\
\hline
ja & 25.00\% \\
en & 13.75\% \\
it & 13.75\% \\
hi & 12.50\% \\
fr & 5.00\% \\
de & 3.75\% \\
zh & 1.25\% \\
\hline
\end{tabular}
\caption{Language distribution of two-component disjoint intervals.}
\label{tab:disjoint-bimodality-language}
\end{table}

\section{Mappers Hyperparameter Tuning and Evaluation}
\label{sec:mappers}

We first outline the final selected hyperparameters used to compare the Similarity, Encoder, and LLM baselines in Table \ref{tab:mapper_final_configs}.
To arrive at the configurations above, the Similarity and Encoder methods underwent hyperparameter tuning. Table~\ref{tab:mapper_search_space} details the search spaces explored.
We tuned mapper hyperparameters with repeated cross-validation. Unless otherwise noted, we used:
\begin{itemize}
\item 4-fold cross-validation;
\item seeds $\{7,13,42\}$;
\item objective $\mathrm{MAE}_{\mu} + \mathrm{MAE}_{\sigma}$.
\end{itemize}

Finally, to quantify the error introduced by the similarity mapper, we compare calibration under two confidence sources for the LC method: (i)~gold human-annotated confidence distributions and (ii)~similarity-mapper predicted confidence distributions.
\begin{itemize}
    \item We use the 7-language annotated set.
    \item For the mapper, we use the best per-language similarity configuration.
    \item Evaluation is performed with 4-fold CV and 3 seeds (7, 13, 42), yielding 12 test splits per language.
    \item Per language: 240 OOF datapoints (80 expressions $\times$ 3 seeds).
    \item DistECE is computed.
\end{itemize}

We use the gold-only convention where labels are treated as correct for all datapoints, and compare:
\begin{itemize}
    \item \textbf{Gold ECE:} using gold confidence.
    \item \textbf{Mapper ECE:} using mapper-predicted confidence.
\end{itemize}

Overall, gold distributions yield an ECE of 0.493, compared to 0.476 when using the similarity mapper ($\Delta_{\text{mapper}-\text{gold}} = -0.016$). 

Additionally, we performed validation to address whether the mapper may be smoothing away cross-linguistic variance.
At the expression level, we compared the human and mapper confidence distributions by measuring Jensen-Shannon divergence. The median divergence was 0.076, suggesting the mapper usually stays close to the human distribution.
We tested whether certainty-level distinctions are preserved within each language with a pairwise Mann-Whitney test with BH correction. Across certainty levels, the mapper preserved 120/125 significant contrasts.

\begin{figure*}
    \centering
    \includegraphics[width=1\linewidth]{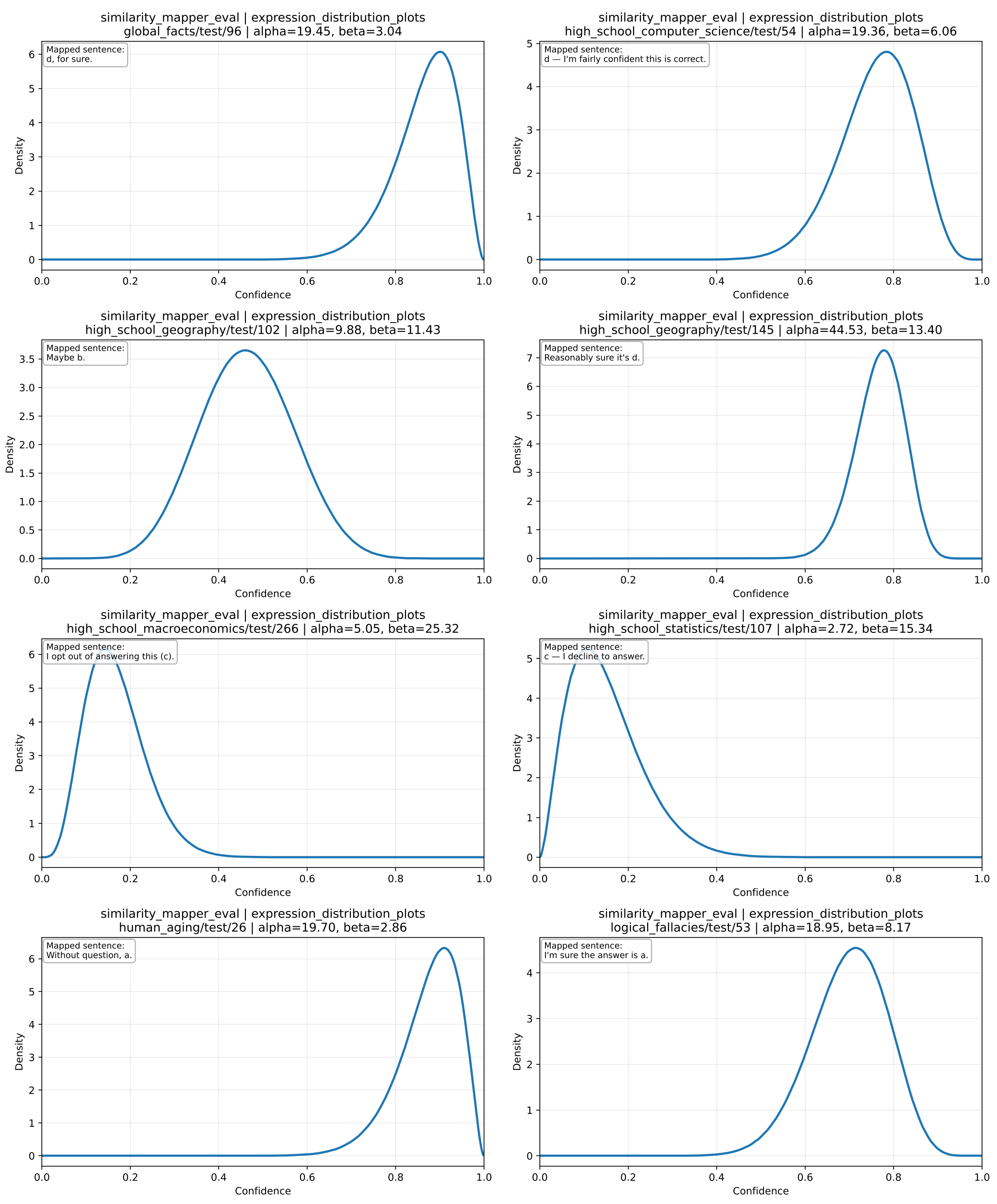}
    \caption{Examples of predicted Beta distributions generated by the similarity mapper for various Qwen responses, showcasing how natural language expressions are translated into continuous uncertainty representations.}
    \label{fig:qwen}
\end{figure*}

Per-language ECE values are reported in Table~\ref{tab:mapper_error_ece}. Figure \ref{fig:qwen}, instead, shows examples of the distributions generated by the mapper for new unseen generations.

\begin{table}[htb]
\footnotesize
    \centering
    \begin{tabular}{lcccll}
        \toprule
        \textbf{Scope} & \textbf{MAE$_{\text{sum}}$ ($\downarrow$)} & \textbf{SD} &\textbf{95\% CI}  & \textbf{$k$} &\textbf{temp} \\
        \midrule
        ja & 0.1060 & 0.0251 &$\pm$0.0142  & 7 &0.03 \\
        fr & 0.1084 & 0.0087 &$\pm$0.0049  & 3 &0.03 \\
        en & 0.1120 & 0.0124 &$\pm$0.0070  & 5 &0.03 \\
        hi & 0.1135 & 0.0057 &$\pm$0.0032  & 7 &0.03 \\
        de & 0.1169 & 0.0133 &$\pm$0.0075  & 7 &0.03 \\
        zh & 0.1181 & 0.0145 &$\pm$0.0082  & 7 &0.03 \\
        it & 0.1249 & 0.0123 &$\pm$0.0069  & 7 &0.03 \\
        \midrule
        all\_lang& 0.1156 & 0.0078 &$\pm$0.0044  & 7 &0.03 \\
        \bottomrule
    \end{tabular}
    \caption{Performance of best similarity mapper across runs (MAE$_{\text{sum}}$, lower is better).}
    \label{tab:primary_performance}
\end{table}

\begin{table}[htb]
\footnotesize
    \centering
    \begin{tabular}{lcc}
        \toprule
        \textbf{Language} & \textbf{Gold ECE} & \textbf{Mapper ECE} \\
        \midrule
        de & 0.514 & 0.500 \\
        en & 0.555 & 0.534 \\
        fr & 0.502 & 0.489 \\
        hi & 0.464 & 0.463 \\
        it & 0.420 & 0.381 \\
        ja & 0.529 & 0.518 \\
        zh & 0.465 & 0.445 \\
        \bottomrule
    \end{tabular}
    \caption{Comparison of Expected Calibration Error (ECE) using gold human-annotated confidence versus similarity-mapper predicted confidence across seven languages.}
    \label{tab:mapper_error_ece}
\end{table}
\begin{table}[htb]
\centering
\resizebox{0.4\textwidth}{!}{
\begin{tabular}{lcccc}
\toprule
\textbf{Language} & \textbf{MAE $\mu$} & \textbf{MAE $\sigma$} & \textbf{RMSE $\mu$} & \textbf{RMSE $\sigma$} \\
\midrule
All & 0.075 & 0.040 & 0.103 & 0.051 \\
en& 0.075 & 0.038 & 0.096 & 0.047 \\
de& 0.080 & 0.037 & 0.110 & 0.046 \\
fr& 0.074 & 0.034 & 0.100 & 0.045 \\
hi& 0.063 & 0.050 & 0.078 & 0.062 \\
it& 0.086 & 0.039 & 0.121 & 0.051 \\
ja& 0.066 & 0.040 & 0.094 & 0.056 \\
zh& 0.080 & 0.038 & 0.112 & 0.048 \\
\bottomrule
\end{tabular}
}
\caption{Performance of the Similarity Mapper Across Languages}
\label{tab:similarity_languages}
\end{table}

\begin{table}[htb]
\centering
\footnotesize
\begin{tabular}{p{0.2\linewidth}p{0.70\linewidth}}
\toprule
\textbf{Mapper} & \textbf{Final Selected Setting} \\
\midrule
\textbf{Similarity} \newline (one-head) & 
    \textbullet~ \texttt{model\_name}: \texttt{intfloat/} \newline \texttt{multilingual-e5-large-instruct} \newline
    \textbullet~ \texttt{k}: 7 \newline
    \textbullet~ \texttt{temperature}: 0.03 \newline
    \textbullet~ \texttt{use\_instruction}: \texttt{true} \newline
    \textbullet~ \texttt{mask\_letters}: \texttt{true} \\
\midrule
\textbf{Similarity} \newline (two-head) & 
    \textbullet~ \texttt{model\_name}: \texttt{intfloat/}\newline\texttt{multilingual-e5-large-instruct} \newline
    \textbullet~ \texttt{k\_mean}: 7, \texttt{k\_std}: 3 \newline
    \textbullet~ \texttt{temperature\_mean}: 0.08, \newline \texttt{temperature\_std}: 0.12 \newline
    \textbullet~ \texttt{aggregation\_mean}: \texttt{weighted\_mean} \newline
    \textbullet~ \texttt{aggregation\_std}: \texttt{weighted\_median} \newline
    \textbullet~ \texttt{std\_calibration\_slope}: 1.1 (other terms at defaults) \newline
    \textbullet~ \texttt{mask\_letters}: \texttt{true} \\
\midrule
\textbf{Encoder} & 
    \textbullet~ \texttt{model\_name}: \texttt{distilbert-base}\newline\texttt{-multilingual-cased} \newline
    \textbullet~ \texttt{max\_length}: 128, \texttt{batch\_size}: 8 \newline
    \textbullet~ \texttt{learning\_rate}: 5e-5, \newline\texttt{num\_train\_epochs}: 8 \newline
    \textbullet~ \texttt{weight\_decay}: 0.05, \texttt{eval\_ratio}: 0.2 \newline
    \textbullet~ \texttt{mask\_letters}: \texttt{true} \\
\midrule
\textbf{LLM Mapper} \newline (finetuned baseline) & 
    \textbullet~ \texttt{model\_name}: \texttt{Llama-3-8B-Instruct} \newline
    \textbullet~ \textbf{Validation}: 5-fold CV with seeds $\{7, 13, 42\}$ \newline
    \textbullet~ \texttt{batch\_size}: 1, \texttt{learning\_rate}: 2e-4, \texttt{epochs}: 3 \newline
    \textbullet~ \texttt{mask\_letters}: \texttt{true} \newline
    \textbullet~ \textbf{LoRA}: $r=16$, $\alpha=32$, dropout $=0.05$ \newline
    \textbullet~ \textbf{Target Modules}: \{\texttt{q\_proj, k\_proj, v\_proj, o\_proj}\} \\
\bottomrule
\end{tabular}
\caption{Best/final mapper settings used for reported experiments.}
\label{tab:mapper_final_configs}
\end{table}
\begin{table}[h]
\centering
\footnotesize
\begin{tabular}{p{0.2\linewidth}p{0.68\linewidth}}
\toprule
\textbf{Mapper} & \textbf{Hyperparameter Search Space} \\
\midrule
\textbf{Similarity} \newline (one-head) & 
    \textbullet~ \texttt{model\_name}: \{\texttt{intfloat/}\newline\texttt{multilingual-e5-large-instruct}\} \newline
    \textbullet~ \texttt{k}: $\{3, 5, 7, 9, 11, 15\}$ \newline
    \textbullet~ \texttt{temperature}: $\{0.03, 0.05,$\newline$ 0.08, 0.12, 0.2\}$ \newline
    \textbullet~ \textbf{Fixed}: \texttt{use\_instruction}=\texttt{true}, \newline\texttt{mask\_letters}=\texttt{true}\\
\midrule
\textbf{Similarity} \newline (two-head) & 
    \textbullet~ \textbf{Search Space}: Two-head knobs \newline including \texttt{k\_mean}, \texttt{k\_std},\newline \texttt{temperature\_mean}, \texttt{temperature\_std}, \texttt{aggregation\_mean/std}, and linear calibration slopes/intercepts. \\
\midrule
\textbf{Encoder} & 
    \textbullet~ \texttt{model\_name}: \{\texttt{xlm-roberta-base}, \newline\texttt{distilbert-base-multilingual-cased}, \newline \texttt{bert-base-multilingual-cased}\} \newline
    \textbullet~ \texttt{max\_length}: $\{128, 192, 256\}$ \newline
    \textbullet~ \texttt{batch\_size}: $\{8, 16\}$ \newline
    \textbullet~ \texttt{learning\_rate}: \{1e-5, 2e-5, 3e-5, 5e-5\} \newline
    \textbullet~ \texttt{num\_train\_epochs}: $\{4, 6, 8\}$ \newline
    \textbullet~ \texttt{weight\_decay}: $\{0.0, 0.01, 0.05\}$ \newline
    \textbullet~ \textbf{Fixed}: \texttt{eval\_ratio}=0.2, \newline\texttt{mask\_letters}=\texttt{true}\\
\bottomrule
\end{tabular}
\caption{Hyperparameter search spaces used for mapper tuning.}
\label{tab:mapper_search_space}
\end{table}

\clearpage
\section{Prompts}
\label{sec:prompts}
In this section, we report examples of the prompts used for confidence elicitation. For each language, the model was prompted in the target language.

\begin{tcolorbox}[colback=gray!5!white,colframe=gray!75!black,title=\textbf{Example Verbalised Numerical Confidence Prompt (English)}, arc=2mm, left=2mm, right=2mm, top=2mm, bottom=2mm]
Give your answer (A/B/C/D) and your confidence as a percentage (0-100\%). Do not give any explanations. 
\\
What is the capital city of Australia?\\
A. Sydney\\
B. Melbourne\\
C. Canberra\\
D. Perth\\

Format:\\
Answer: <letter>\\
Confidence: <number>

Answer: 
\end{tcolorbox}

\begin{tcolorbox}[colback=gray!5!white,colframe=gray!75!black,title=\textbf{Example Linguistic Confidence Prompt (English)}, arc=2mm, left=2mm, right=2mm, top=2mm, bottom=2mm]
Give your answer (A/B/C/D) and describe your confidence using a linguistic hedge. Use the following format.\\

What is the capital city of Australia?\\
A. Sydney\\
B. Melbourne\\
C. Canberra\\
D. Perth\\

Format:\\
Answer: <letter>\\
Confidence: <hedge phrase>\\

Answer:
\end{tcolorbox}

\begin{tcolorbox}[colback=gray!5!white,colframe=gray!75!black,title=\textbf{Example Linguistic Confidence with ICL Prompt (English, 3/5 ICL examples displayed)}, arc=2mm, left=2mm, right=2mm, top=2mm, bottom=2mm]
Give your answer (A/B/C/D) and describe your confidence using a linguistic hedge. Use the following format.\\

Which of these is commonly used to treat allergies?\\
  A. antimatter\\
  B. anticoagulants\\
  C. antiseptics\\
  D. antihistamines\\
Answer: D\\
Confidence: d with very little confidence.\\
\\
Which of the following is true for all bases?\\
  A. All bases donate OH- ions into solution.\\
  B. Only strong bases create solutions in which OH- ions are present.\\
  C. Only strong bases are good conductors when dissolved in solution.\\
  D. For weak bases, the concentration of the OH-ions exceeds the concentration of the base in the solution.\\
Answer: C\\
Confidence: Possibly c.\\
\\
What is the optimal span of control?\\
  A. 2\\
  B. 5\\
  C. 7\\
  D. None of the above\\
Answer: D\\
Confidence: d — I firmly believe this is the right answer.\\
\\
What is the capital city of Australia?\\
A. Sydney\\
B. Melbourne\\
C. Canberra\\
D. Perth\\

Format:\\
Answer: <letter>\\
Confidence: <hedge phrase>

Answer:
\end{tcolorbox}

\begin{tcolorbox}[colback=gray!5!white,colframe=gray!75!black,title=\textbf{Example Base Prompt for Logits- and Multisampling-based methods (English)}, arc=2mm, left=2mm, right=2mm, top=2mm, bottom=2mm]
Answer the following multiple-choice question. Respond with only the letter (A, B, C, or D).
\\

What is the capital city of Australia?\\
A. Sydney\\
B. Melbourne\\
C. Canberra\\
D. Perth\\

Answer:
\end{tcolorbox}

%\newpage
\section{Supplementary Results}
\label{sec:results}

\subsection{Validation on Open Question-Answering}
\label{sec:openqa}
We validate calibration for linguistic confidence on a sample of GlobalMMLU Lite without presenting the models with the answer options. In particular, we randomly sample 500 question for each of the 7 languages supported by our dataset and mapper. We prompt a medium-sized instruct model per model family analysed: aya-expanse-8b, gemma-3-12b-it, llama-3.1-8B-Instruct, Qwen3.5-9B, and EuroLLM-9B-Instruct.
We compare the two following confidence elicitation methods:
\begin{itemize}
    \item Linguistic Confidence in the zero shot setting;
    \item For the logit-based method, as we do not have option answers, we look at the mean probability of generated answer tokens, a commonly used simple approach \cite{geng2024survey, vazhentsev2023hybrid}.
\end{itemize}
As the answers cannot be evaluated using exact matching to the target, we employ Gemini-3.5-flash as a judge, with the following prompt. After running the evaluator 3 times, the self-agreement rate is 97.22\%. Additionally, the comparison between LLM-as-a-judge with Gemini-3.5-flash and token-overlap F1 between target and generated answers shows a Pearson correlation of 0.4. 
As the results in Table \ref{tab:openqa} show, in open-ended generation, linguistic calibration performs better than token probabilities; mean token probabilities track fluency rather than factual correctness, and factors such as tokenization variations, answer length, and the high probability assigned to common filler tokens are likely to introduce noise.
It’s important to note that, in the MCQ setting, logit-based confidence serves as a valid probe for internal model certainty because the confidence target (the probability of the correct option) is aligned with the decision target (the single A/B/C/D token being generated).
Regardless of the task setting, there is a persistent mismatch between internal and verbalised confidence and cross-linguistic performance gaps. The fact that elicitation methods’ assigned confidence scores diverge reinforces the need for more reliable, multilingual, calibration methods like better linguistic confidence.

\begin{tcolorbox}[colback=gray!5!white,colframe=gray!75!black,title=\textbf{LLM-as-a-judge}, arc=2mm, left=2mm, right=2mm, top=2mm, bottom=2mm]
You are a strict but fair multilingual QA evaluator.
Task: judge whether the model prediction is correct given the references.
Rules:\\
1) Correct if prediction meaning is entailed by at least one reference.\\
2) Ignore minor wording differences, punctuation, and formatting.\\
3) Mark incorrect if prediction contradicts reference facts, adds wrong specifics, or is unsupported.\\
4) Return ONLY valid JSON with keys: correct (bool), score (0..1), reason (short string).\\
5) score reflects semantic correctness confidence, not style quality.\\
\\
Language: {lang}\\
Question: {question}\\
References:\textbackslash{refs\_text}\\
Prediction:\textbackslash{prediction}\\
Return JSON only.

\end{tcolorbox}

\begin{table}[h!]
\footnotesize
    \centering
    \begin{tabular}{>{\raggedright\arraybackslash}p{0.15\linewidth}ccll}
        \toprule
        & \textbf{ECE}& \textbf{Brier} & \textbf{AUROC}& \textbf{Acc.}\\
        \midrule
        Mean token prob.& 0.537&  0.489& 0.554& 0.277\\
        LC& 0.490&  0.405& 0.564& 0.251\\
    \end{tabular}
    \caption{Comparison of Expected Calibration Error (ECE), brier score, AUROC, and Accuracy in the free-form generation setting}
    \label{tab:openqa}
\end{table}

\subsection{Statistical Analysis of Cross-linguistic Performance}
\label{sec:stat}
\begin{figure*}[b]
        \centering
        \includegraphics[width=1\linewidth]{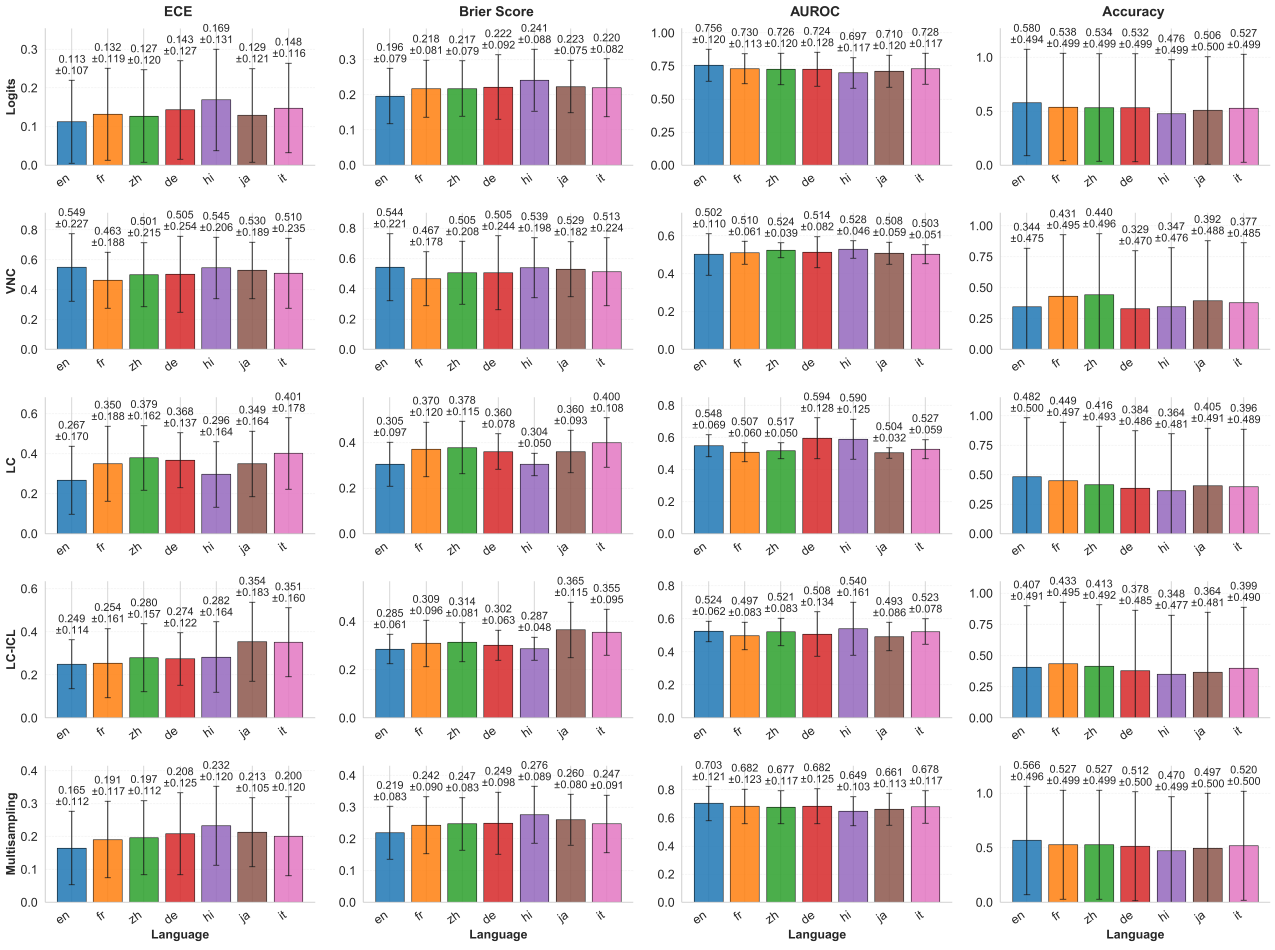}
        \caption{Detailed breakdown of performance metrics (ECE, Brier Score, AUROC, and Accuracy) across languages and confidence elicitation method. The gap in performance, calibration, and discriminability between languages varies depending on the elicitation method: while the logit-based and multisampling methods show narrower gaps, handling multilingual data with the most stability, the verbalised methods are less predictable and subject to higher sensitivity to the language used.}
        \label{fig:lang}
\end{figure*}

For every pair of languages (i.e. En vs. It), every pair of methods (i.e. Logit vs LC), and every pair of models, we conducted a Wilcoxon signed-rank test. This non-parametric test can be used for paired data to determine if there are significant differences between the distributions of the performance metrics.
To account for the likelihood of false positives arising from a large number of pairwise comparisons, we applied the Benjamini-Hochberg (FDR) correction to calculate q-values/adjusted p-values.
A difference is considered statistically significant only if q<0.05.
Overall, we find that:
    \begin{itemize}
        \item With respect to Calibration, Brier score, and accuracy, English performs significantly better than all other languages apart from French, where the difference is not significant (ECE: q=0.122, accuracy: q=0.058). With respect to AUROC, fewer pairs are significant, and English performs significantly better than French (q=0.002), Italian (q=0.028), Japanese (q=4.16e-07), and Chinese (q=0.011).
        \item French achieves significantly better ECE than all languages apart from English and Chinese, where the differences are not significant.
        \item Chinese is significantly better calibrated than Japanese (q=0.026) and Italian (q=0.006).
        \item German is significantly better calibrated than Italian (q=0.025).
        \item The sensitivity to language variation depends on the elicitation method. Logit and multisampling tend to have smaller but consistent language gaps (with 46/84 significant language pairs across metrics for the logit-based method), while verbalised methods are more heterogeneous, with less consistent language performance. LC-ICL shows 16/84 significant pairs, LC shows 42/84, and VNC 10/84; as variance is larger, test statistic weakens. Figure \ref{fig:lang} shows the per-language performance for each elicitation method.
        \item The performance gap between English and other languages is present across model families, but we do not identify any significant differences between them, indicating limited within-model power.
    \end{itemize}

To assess whether the cross-linguistic gaps we report are an artifact of differing task accuracy, we conducted two complementary analysis:

\begin{enumerate}
    \item We look at Spearman’s rank correlations between accuracy and each metric to confirm local monotonic trends. Even after applying Benjamini-Hochberg (FDR) correction, higher task accuracy is significantly correlated with lower calibration error (ECE and Brier score).
    \item To isolate whether calibration and discriminability gaps are artifacts of raw correctness, we fit a model without accuracy and a model with accuracy, both including language/model/method as categorical predictors. Because language is a categorical predictor, we fix English as the baseline to compare other languages to, and each coefficient reported below is to be interpreted as a language’s expected metric value difference from English, all else (i.e. accuracy level, method, model) being equal.
\end{enumerate}

We fit a baseline model ($Metric \sim C(Language) + C(Model) + C(Method)$) and an accuracy-adjusted model ($Metric \sim Accuracy + C(Language) + C(Model) + C(Method)$).
Comparing these two regression models lets us separate the total language effect into the accuracy-explained component (the reduction in effect magnitude after controlling for accuracy), and the residual language component (the remaining effect after accounting for accuracy).

\begin{itemize}
    \item For ECE and Brier scores, accuracy adds a strong explanatory power ($\Delta R^2$ = 0.133 for ECE, 0.103 for Brier). Higher accuracy correlates with lower calibration error. However, accuracy does not explain the entire cross-linguistic performance gap: the regression analysis shows that language is a significant independent predictor of calibration error (ECE: p = 2.862e-7, Brier: p = 1.818e-4) even after adjusting for accuracy level.
    \item The role of accuracy when evaluating AUROC is not statistically significant (p = 0.347, $\Delta R^2$ = 4.097e-5), and the residual effect is also weaker.
    \item The influence of language and accuracy is not consistent across methods. We find that VNC, where the model is asked to produce a numerical score for its uncertainty, shows the highest accuracy-driven behaviour ($\Delta R^2$ = 0.172 for ECE; 0.169 for Brier), while language is a significant factor in performance for zero-shot Linguistic Confidence elicitation (ECE: p=2.73e-14, Brier: p=1.28e-12).
\end{itemize}

\subsection{Discrete Confidence Scores vs. Confidence Distributions}

\begin{figure*}
    \centering
    \includegraphics[width=1\linewidth]{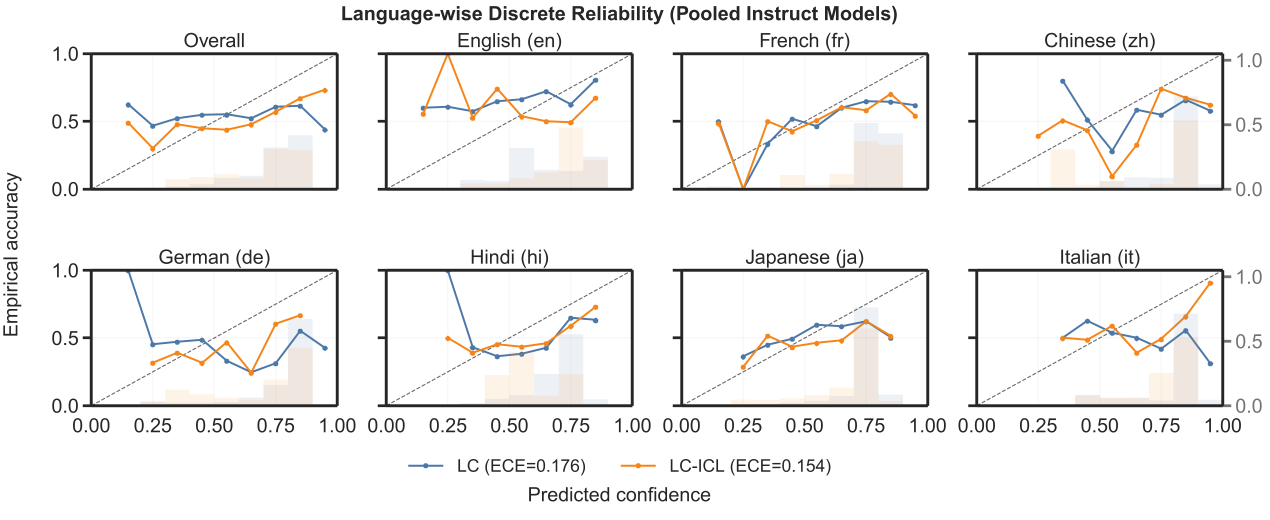}
    \caption{Language-wise discrete reliability across pooled instruct models (10 uniform bins). Solid lines indicate the empirical accuracy within each confidence bin, while the faded background bars represent the frequency of predictions in each bin. The dashed diagonal line denotes perfect calibration. Languages like German (de), Hindi (hi), and Chinese (zh) display jagged U-shaped curves, suggesting severe miscalibration in moderate confidence ranges. However, these sharp discontinuities are largely binning artifacts; forcing continuous probabilities into uniform boxes results in exceptionally small sample sizes in lower-confidence bins, making discrete ECE a highly noisy, high-variance estimator in these regions.}
    \label{fig:disc}
\end{figure*}

\begin{figure*}
    \centering
    \includegraphics[width=1\linewidth]{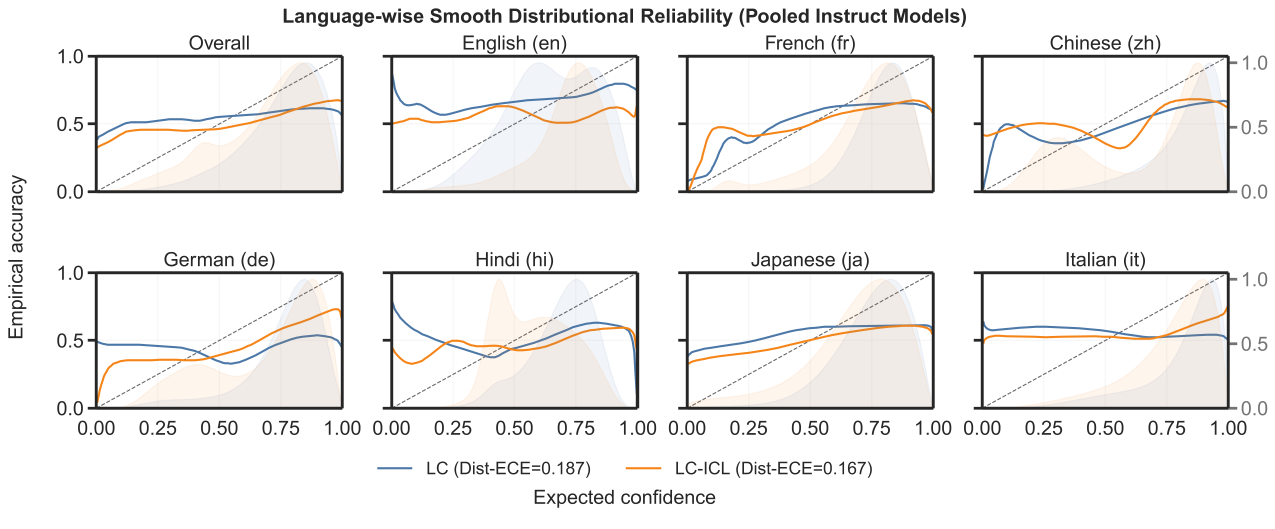}
    \caption{Language-wise smooth distributional reliability across pooled instruct models. Solid lines map the continuous, smoothed empirical accuracy against expected confidence, while the shaded regions represent the normalised prediction density across the probability space. Unlike discrete binning, this kernel-smoothed approach handles data sparsity and shows much more stable, monotonic calibration trends. The shaded density maps highlight that the vast majority of prediction mass is concentrated in high-confidence regions (\~0.7–0.9) across most languages. By incorporating density and eliminating arbitrary bin boundaries, this metric provides a more statistically robust assessment.}
    \label{fig:dist}
\end{figure*}

This section shows the evaluation of model calibration using both discrete ECE in Figure \ref{fig:disc} and treating hedges as confidence distributions in Figure \ref{fig:dist}, following \citet{wang2025calibratingexpressionscertainty}. We assess the performance of pooled instruct models across eight languages.

\clearpage

\end{document}